\documentclass[10pt,journal,cspaper,compsoc]{IEEEtran}
\usepackage[breaklinks=true,colorlinks]{hyperref}

\usepackage[nocompress]{cite}
\usepackage{algorithmic}
\usepackage{subfigure}
\usepackage{url}
\usepackage{multicol,multirow}
\usepackage[ruled,vlined]{algorithm2e} 
\usepackage{times}

\usepackage{amsmath}
\usepackage{amssymb}
\usepackage{ragged2e}

\usepackage{overpic}
\usepackage{rotating}
\usepackage{color}
\usepackage{xcolor}

\usepackage{etoolbox}
\makeatletter \patchcmd{\@makecaption}{\\}{.\ \justifying }{}{}\makeatother
	
\usepackage{graphicx}
\DeclareGraphicsExtensions{.pdf,.jpg,.png}

\def\eg{\emph{e.g.}}
\def\etc{\emph{etc}}
\def\etal{{\em et al.~}}
\def\sArt{{state-of-the-art~}}
\def\coca{\emph{CoCA}~\cite{zhang2020gicd}~}

\newcommand{\revise}[1]{\textcolor{black}{#1}}
\newcommand{\mathvec}[1]{\boldsymbol{#1}}
\newcommand{\mathmat}[1]{\mathbf{#1}}
\newcommand{\mathset}[1]{\mathcal{#1}}

\renewcommand{\tablename}{Table }

\newcommand{\figref}[1]{Fig.~\ref{#1}}
\newcommand{\tabref}[1]{Tab.~\ref{#1}}
\newcommand{\secref}[1]{Sec.~\ref{#1}}
\renewcommand{\eqref}[1]{Eq.~\ref{#1}}
\newcommand{\algref}[1]{Algorithm \ref{#1}}

\newcommand{\myPara}[1]{\vspace{10pt}\noindent\textbf{#1.}\quad}

\newcommand{\tablestyle}[2]{\setlength{\tabcolsep}{#1}\renewcommand{\arraystretch}{#2}\centering}

\RequirePackage{silence}
\graphicspath{{./figures/}}

\begin{document}

%%%%%%%%% TITLE
\title{Co-Salient Object Detection with Co-Representation Purification}
\author{Ziyue Zhu* \quad Zhao Zhang* \quad Zheng Lin \quad Xing Sun \quad Ming-Ming Cheng
\IEEEcompsocitemizethanks{
\vspace{-5pt}
\IEEEcompsocthanksitem Z. Zhu, Z. Zhang, Z. Lin, and M.M. Cheng are with
the TKLNDST, College of Computer Science, Nankai University, 
Tianjin 300350, China.
\IEEEcompsocthanksitem Z. Zhang is with SenseTime Research.
\IEEEcompsocthanksitem X. Sun is with Youtu Lab, Tencent.
\IEEEcompsocthanksitem * denotes equal contribution.
\IEEEcompsocthanksitem M.M. Cheng is the corresponding author (cmm@nankai.edu.cn).
}% <-this % stops an unwanted space
}

%%%%%%%%% ABSTRACT
\IEEEcompsoctitleabstractindextext{%
\begin{abstract}
\justifying
Co-salient object detection (Co-SOD) aims at discovering the common objects 
in a group of relevant images.
Mining a co-representation is essential for locating co-salient objects.
Unfortunately, the current Co-SOD method does not pay enough attention that 
the information not related to the co-salient object is included 
in the co-representation.
Such irrelevant information in the co-representation interferes 
with its locating of co-salient objects.
In this paper, we propose a Co-Representation Purification (CoRP) method 
aiming at searching noise-free co-representation.
We search a few pixel-wise embeddings probably belonging to co-salient regions.
These embeddings constitute our co-representation and guide our prediction.
For obtaining purer co-representation, we use the prediction to 
iteratively reduce irrelevant embeddings in our co-representation.  
Experiments on three datasets demonstrate that our CoRP achieves
\sArt performances on the benchmark datasets.
Our source code is available at \href{CoRP}{https://github.com/ZZY816/CoRP}.
\end{abstract}

}

\maketitle
\IEEEdisplaynotcompsoctitleabstractindextext
\IEEEpeerreviewmaketitle

%%%%%%%%% BODY TEXT
\section{Introduction} \label{intro}

\begin{figure}[t]
	\centering
	\small
	\begin{overpic}[width=1\columnwidth]{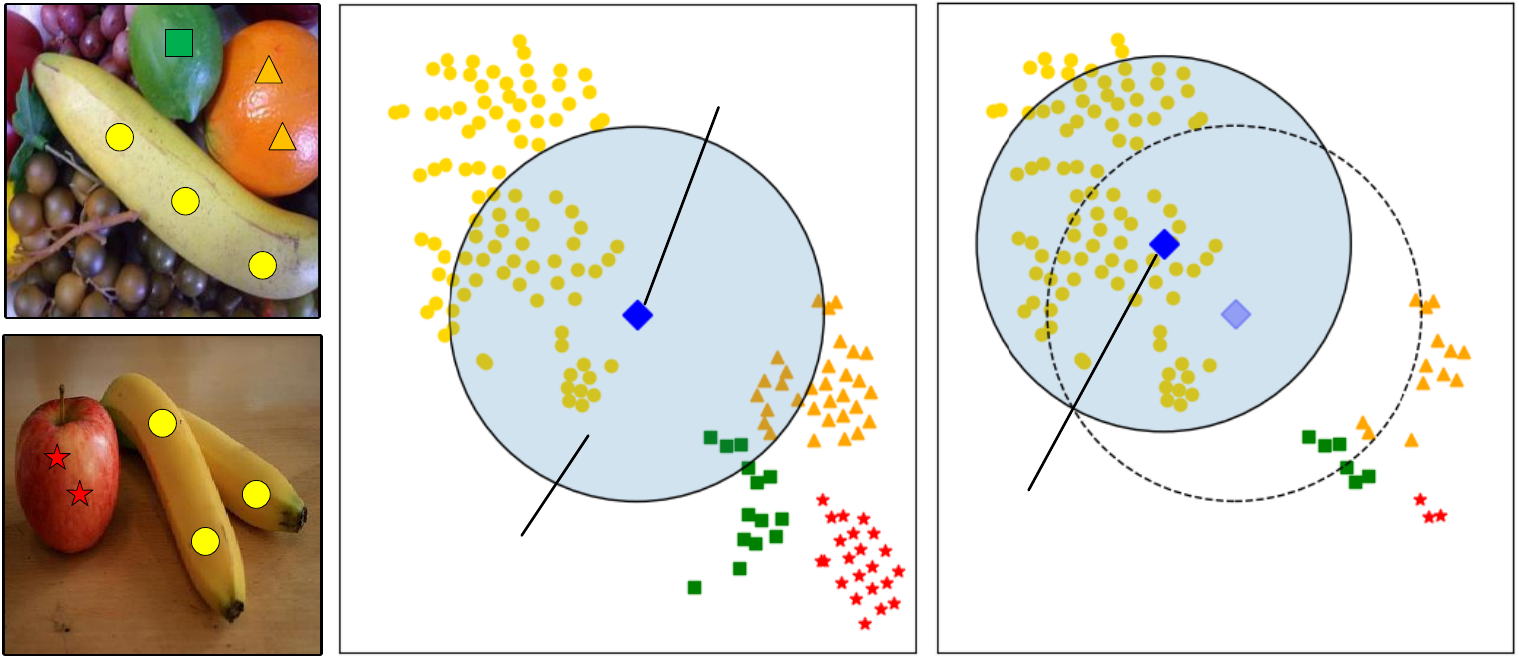}
	\put(38, -3.5){(a)}
	\put(78, -3.5){(b)}
	\put(23, 5.5){{co-representation}}
	\put(41, 38){{initial center}}
	\put(63, 8){{new center}}
	\end{overpic}
	\caption{ \textbf{t-SNE~\cite{van2008tsne} visualization of embeddings.}
	 (a) ``\textcolor{yellow}{$\bullet$}'' represent embeddings of 
	 co-salient objects ``banana''.
	 We observe that the embeddings (in the blue circle area) quite near 
	 the center are very likely to belong to co-salient objects. 
	 We employ them as our co-representation to localize co-salient objects. 
	 (b) When many embeddings of irrelevant objects are filtered out 
	 by our primary prediction, 
	 we can obtain a new center less interfered by irrelevant embeddings. 
	 The new center helps search purer co-representation leading to 
	 more accurate prediction.
  }\vspace{-10pt}
	\label{fig:main}
\end{figure}

\IEEEPARstart{H}{u}man perception system~\cite{cong2018review} 
can effortlessly discover the most salient area. %visual attactive
Co-salient object detection (Co-SOD) aims at 
discovering common salient objects from a group of relevant images.
Meanwhile, Co-SOD needs to deal with unseen object categories,
which are not learned during the training process.
Such ability can serve as the preprocessing step for 
many real-world applications, \eg, video co-localization 
\cite{Jerripothula2016videoco, Joulin2014co-lofrank}, 
semantic segmentation~\cite{yu2019semanticseg}, 
image quality assessment~\cite{wang2019assessment}, 
and weakly supervised learning~\cite{Zhang2019boost}.
The difficulty of the Co-SOD task lies in discovering co-salient objects 
in the context of cluttered real-world environments.
As shown in \figref{fig:main}, 
it is challenging to automatically discover and segment 
the co-salient objects ``banana'' among multiple irrelevant salient objects.

To distinguish co-salient objects, most \sArt (SOTA) methods
\emph{directly} estimate a co-representation 
to capture the shared characteristics of co-salient objects,
by feature aggregation \cite{Tsai2019opt,wei2019group-wise},
clustering~\cite{zha2020robust,Zhang2020GCAGC}, 
principal component analysis \cite{zhang2019csmg,deng2021re},
global pooling~\cite{zhang2020gicd,fan2021GCoNet,Jin2020ICNet}, \etc. 
The co-representations of these methods are summarized from all regions
\cite{zhang2020gicd,zha2020robust, zhang2019csmg}, 
or within pre-predicted salient regions 
\cite{Jin2020ICNet,zhang2020CoADNet}.
Although achieving promising performances in many scenes,
they often ignore the noisy information related to irrelevant salient objects.

Utilizing noisy co-representation can lead to incorrect localization 
of co-salient objects, limiting Co-SOD models' performances, 
especially for complex real-world scenes. 
To overcome this bottleneck, 
we try to reduce the irrelevant information in the co-representation.
Unlike current methods directly obtaining the co-representation 
by summarizing all regions
\cite{zhang2020gicd, wei2019group-wise,zha2020robust, zhang2019csmg} 
or salient regions~\cite{Jin2020ICNet,zhang2020CoADNet}, 
we propose an iterative process to search confident locations only 
belonging to co-salient regions as our co-representation, 
which guides the complete segmentation of co-salient objects.  
% A question has raised,
% is it possible to take the advantage of the informative but imperfect 
% co-representations and reduce the side-effect of irrelevant information?

Specifically, we propose pure co-representation search (PCS) first to find 
confident embeddings belonging to co-salient regions as our co-representation.
As shown in \figref{fig:main}, among all pixel embeddings of salient objects, 
the embeddings of co-salient objects dominate because of the repetitiveness 
of co-salient objects in the image group.
When obtaining the center by summarizing all the embeddings of salient regions, 
we find that the embeddings closer to the center are more likely to be 
of the co-salient ones. 
Based on this observation, instead of directly using the imperfect center 
to detect co-salient objects~\cite{Jin2020ICNet, deng2021re}, 
we treat the center as a proxy for indexing embeddings that have 
high correlation with it as co-representation. 
Compared with the proxy summarized from all the salient regions, 
our co-representation consisting of confident co-salient embeddings 
is less interfered by irrelevant noise.

\begin{figure*}[ht]
  \centering
	\includegraphics[width=.92\textwidth]{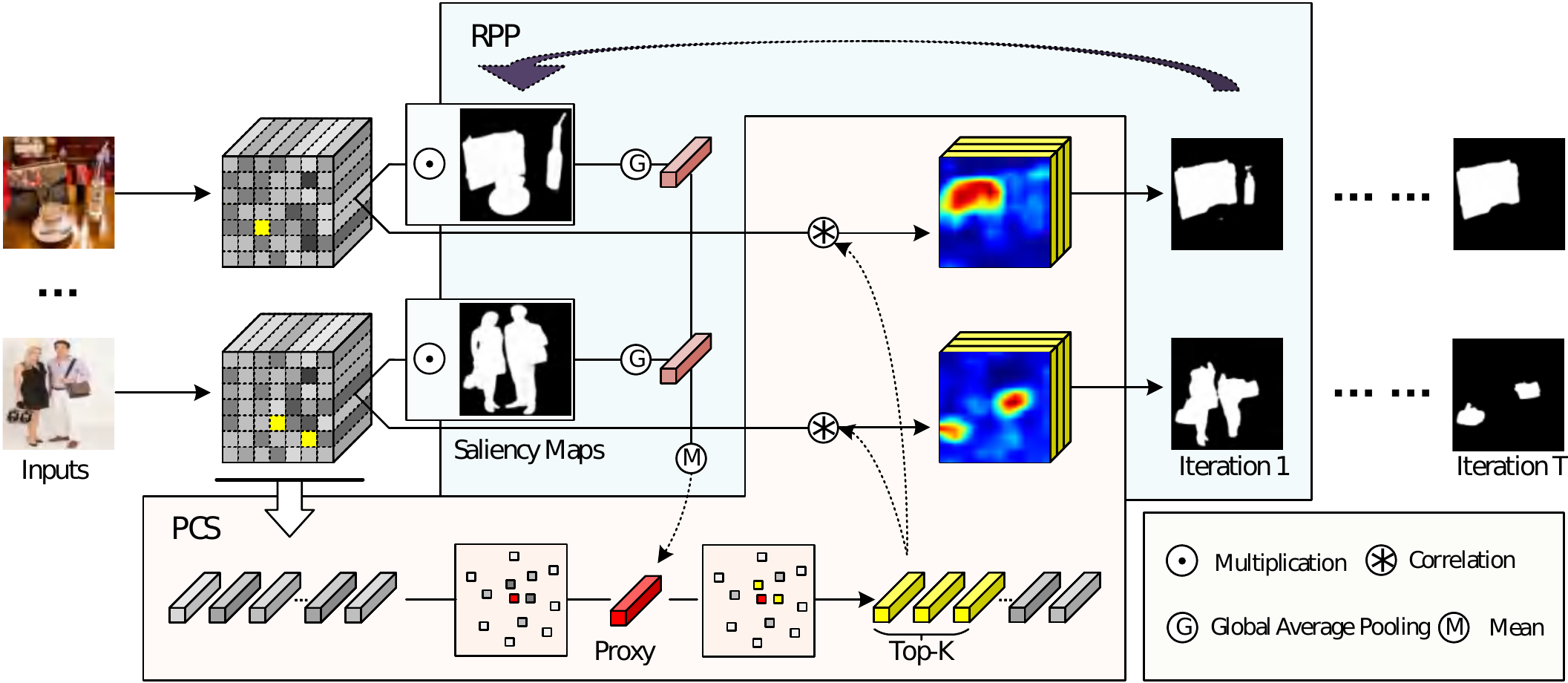}
  \caption{\textbf{Overall framework of our CoRP.} 
    ``PCS" and ``RPP" denote the proposed pure co-representation search 
		(\secref{sec:pcs}) and recurrent proxy purification (\secref{sec:rpp}).
    As shown above, when receiving a group of  images, 
    the corressponding saliency maps are firstly predicted by a 
		backbone-shared saliency object detection (SOD) head.
    A co-representation proxy is generated upon filtering background noise 
		by the saliency maps.
	  With the help of the proxy, PCS searches pure co-representation, 
		which guides co-saliency prediction. 
    RPP feeds back co-saliency maps to calculate a new proxy, 
		which helps searching purer co-representation.
    With the collaboration of PCS and RPP, 
		the noise in the predictions is iteratively removed. 
	  For brevity, we do not draw our encoder-decoder architecture and the SOD head, 
		which shares backbone parameters with this Co-SOD network. 
  }\label{fig:flowchart}
\end{figure*}

Considering that the indexed co-representation from PCS still contains 
irrelevant embeddings,
we propose recurrent proxy purification (RPP),
using predicted co-saliency maps to purify the co-representation iteratively.
After obtaining the prediction of co-salient maps, 
we use the prediction to filter out more noise and acquire a new proxy.
The new proxy helps PCS search the co-representation with less noise 
for more accurate prediction.
We carry out the above process iteratively to purify our co-representation.
Under the alternate work of PCS and RPP, 
the irrelevant embeddings in our co-representation are gradually reduced.
That is, the iteration process makes our representation purer and purer.
We abbreviate our method as CoRP (Co-Representation Purification) 
in the following sections.
In summary, our major contributions are given as follows.

\begin{itemize}
  \item We propose two purification strategies:
		(i) PCS for minning noise-free co-representation 
		% from locations belonging to co-salient objects  
    and (ii) RPP for iteratively reducing noise 
		based on the previous co-saliency maps.
	\item CoRP achieves SOTA performances on challenging datasets,
	\textit{CoCA}~\cite{zhang2020gicd}, 
	\textit{CoSOD3k}~\cite{fan2020taking}, 
	and \textit{CoSal2015}~\cite{zhang2016CoSal}. 
\end{itemize}

\section{Related Works}

\myPara{Co-SOD by low-level consistency}
Since Jacobs~\etal\cite{Jacobs_UIST2010} firstly proposed the task of 
searching common salient objects among a series of relevant images,
Co-SOD has been explored by the community for more than a decade.
Early methods ~\cite{Li2011imagepairs,Jerripothula2016videoco} focus on 
exploring the low-level consistency of co-salient objects in multiple images.
% %
Several methods ~\cite{Tsai2019opt,Chen2010PreattentiveCD} 
extract co-salient objects by gathering similar pixels across 
different images or common low-level cues from feature distributions.
Other studies explore shared cues of a group of images by 
clustering~\cite{Fu2013ClusterCo}, 
metric learning~\cite{Han2017metriclearning}, 
or efficient manifold ranking~\cite{Li2015Efficient}. 
Before searching inter-image repetitiveness, 
some other works~\cite{cao2014self,Li2011imagepairs,chang2011co}
utilize the saliency maps of images to filter background noise to 
more accurately extract the co-salient objects.

Recently, multiple deep learning-based methods
\cite{Hsu_2018_ECCV,Li2019robustrecurrent,zhang2016CoSal} 
have sprung up and significantly outperform previous traditional methods.
These methods mainly focus on learning the co-representation of the 
image group and using it as a constraint signal 
for discovering the co-salient objects.

\myPara{Co-SOD by deep consistency}
Wei~\etal\cite{wei2019group-wise} concatenated a group of image features 
as the co-representation,
and merged it back to each single image feature for automatic prediction.
CSMG~\cite{zhang2019csmg} applies a modified principal component analysis 
algorithm on image features 
to extract the co-representation used to generate prior masks. 
Zha~\etal\cite{zha2020robust} compressed a group of features into a vector 
through the SVM classifier and regarded the vector as co-representation.
GCAGC~\cite{Zhang2020GCAGC} generates co-attention maps through clustering 
and regards the cluster center as the co-representation. 
GICD~\cite{zhang2020gicd} feeds each sample of the group images into a 
pre-trained classifier and sums up all the output as the co-representation. 
GCoNet~\cite{fan2021GCoNet} explores intra-group compactness and 
inter-group separability 
using the co-representation generated by group average pooling.

\myPara{Co-SOD with salient regions refinements} 
Some methods utilize saliency prediction to improve their performances.
CoADNet~\cite{zhang2020CoADNet} designs a group-wise channel shuffling 
for breaking the limit of the fixed input numbers.
The shuffled features masked by saliency maps are concatenated together 
to obtain a co-representation.
ICNet~\cite{Jin2020ICNet} weighted averages each feature masked by 
pre-predicted saliency maps as co-representation, 
which then be used to compare cosine similarity with each position 
for each sample.
CoEGNet~\cite{deng2021re} employs a co-attention projection algorithm 
to extract the co-representation, 
which is utilized to generate co-attentions maps.

Exsiting methods mainly extract co-representations by directly discovering
consistent representation using, 
\eg~clustering~\cite{Fu2013ClusterCo,zha2020robust,Zhang2020GCAGC}, 
principal component analysis \cite{zhang2019csmg,deng2021re},
metric learning~\cite{Han2017metriclearning}, 
manifold ranking~\cite{Li2015Efficient},
or global pooling~\cite{zhang2020gicd,fan2021GCoNet,Jin2020ICNet}. 
Even with salient region refinements, 
it is still difficult for these methods to avoid noisy co-representation
associated with irrelevant salient objects.

\begin{figure*}[ht]    
	\centering
	\begin{overpic}[width=1\textwidth]{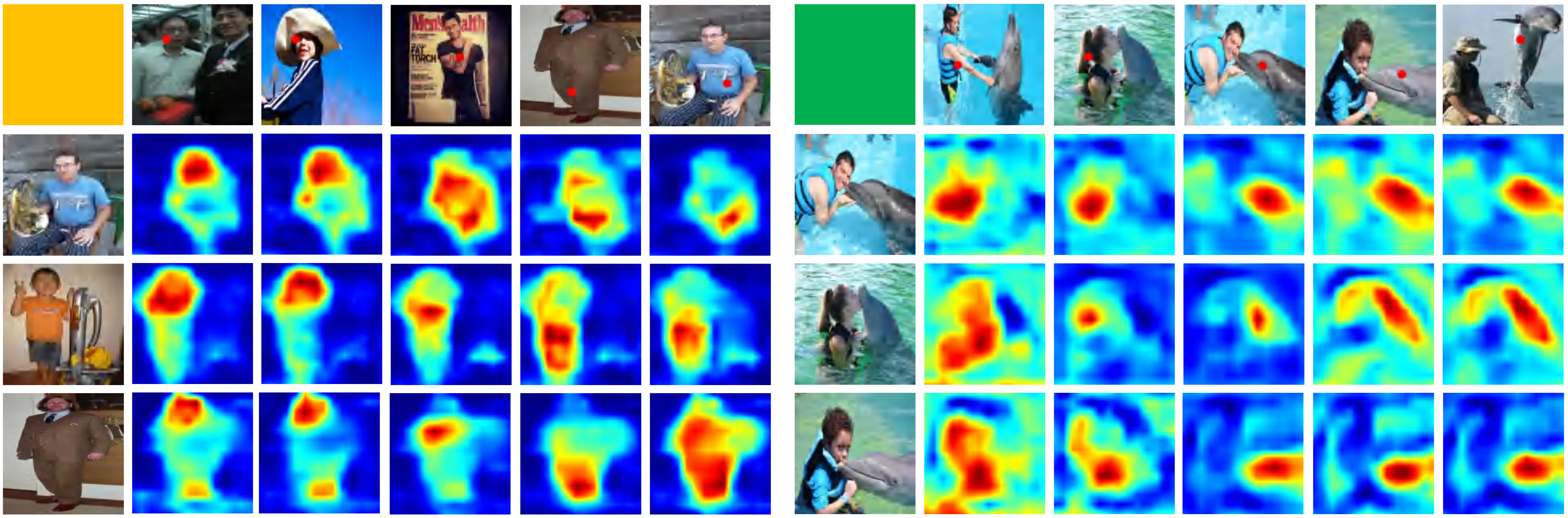}
 	  \put(1.5,28){Person}
 	  \put(51.4,29){Dolphin}
 	  \put(51,27){\&Person}
	\end{overpic}
	\caption{\textbf{Visualization of the locations of embeddings in 
	  our pure co-representation and the channels of feature 
		$\mathmat{A}^t$ transformed by our pure co-representation.}
		In the first row, each of the red point ``\textcolor{red}{$\bullet$}" 
		in one picture represents one of the top-$K$ spatial positions whose 
		corresponding representation constitutes the pure co-representation.
    Each heatmap represents a channel of the converted feature 
		(see \secref{sec:pcs}). 
    It is a correlation map calculated from the deep representation of 
		the target image and the representation noted in the above red point.
	}
	\label{fig:red_hot}
\end{figure*}

\myPara{\revise{Salient Object Detection}}
\revise{Saliency object detection (SOD) aims at discovering regions that 
attract human visual attention in single images \cite{BorjiCVM2019}. 
Traditional SOD methods \cite{itti1998model, cheng2015global, jiang2013salient, 
li2013saliency, perazzi2012saliency} mainly rely on hand-crafted features 
to exploit low-level cues.
Benefiting from the development of CNNs in segmentation tasks, 
many recent SOD methods \cite{wei2020f3net, hou2017deeply, liu2022poolnet+, 
lin2017feature, zhang2018progressive, liu2018picanet} 
design novel network architectures and make pixel-wise predictions.}

\revise{Both SOD and Co-SOD are binary segmentation tasks,
whose ground truth is binary masks.
SOD provides several aspects of benefits for the Co-SOD studies.
Co-SOD methods \cite{Jin2020ICNet, zhang2020CoADNet} can employ single saliency 
maps to filter background noise and better localize co-salient objects.
Meanwhile, the accurate single saliency maps can improve the ability of 
Co-SOD method \cite{deng2021re} to segment object details.
Further, after the jigsaw strategy was proposed by GICD \cite{zhang2020gicd},
a large amount of SOD training data can be utilized to train Co-SOD models.
}

\section{Proposed Method}

\subsection{Overview}

Given a group of images $\mathset{I}=\{\mathmat{I}_n\}_{n=1}^N$,
co-saliency object detection aims to discover their commonly salient object(s) 
denoted by the co-saliency maps $\mathset{M}=\{\mathmat{M}_n\}_{n=1}^N$.
Our CoRP works with two key complementary purification procedures: 
Pure Co-representation Search (PCS) and Recurrent Proxy Purification (RPP).
Our PCS is responsible for estimating co-representation with 
less irrelevant information and locating co-salient objects, 
while RPP utilizes predictions to further purify our co-representation.

As shown in \figref{fig:flowchart},
our PCS starts by extracting $\ell_2$-normalized deep representation 
$\mathset{F}=\{\mathmat{F}_n\in\mathbb{R}^{D\times H \times W}\}_{n=1}^N$ 
for each sample in $\mathset{I}$.
For each of the $N\times H \times W$ spatial positions in the image group,
there is a $D$ dimensional feature vector, 
either corresponding to the target common salient object(s) or 
the irrelevant foreground and background.
Similar to \cite{Jin2020ICNet,deng2021re}, 
we use saliency maps predicted by a backbone-shared SOD head as masks 
to pick feature vectors corresponding to potential common object regions,
which are averaged to obtain a co-representation proxy 
$\mathvec{p}\in\mathbb{R}^{D}$.
While this co-representation proxy contains rich information about the 
co-salient objects, 
the averaged representation $\mathvec{p}$ could easily contain 
irrelevant foreground information.
Thus we try to search for a few most confident pixel-wise embeddings from the 
image features group for representing the common objects.
More specifically, $K$ embeddings 
$\{\mathvec{c}_k\in\mathbb{R}^D\}_{k=1}^{K}$ from $\mathset{F}$
are selected according to their distance to the co-representation proxy 
$\mathvec{p}$
as our co-representation $\mathmat{C}\in\mathbb{R}^{K\times D}$.

For obtaining purer co-representation $\mathmat{C}$, 
RPP utilizes our predicted co-saliency maps to purify our co-representation.
To be specific, RPP feeds back existing prediction $\mathset{M}$ to 
calculate more accurate co-representation proxy $\mathvec{p}$.
With a new co-representation proxy, PCS can explore purer co-representation, 
which finally leads to better prediction.
Our CoRP gradually eliminate the interference of background 
and extraneous foreground noise.
We detail PCS and RPP in \secref{sec:pcs} and \secref{sec:rpp}.

% \begin{figure}[t]
% 	\centering
% \centering
% 	\begin{overpic}[width=0.92\columnwidth]{./figures/red_hot_final.pdf}
% 	\put(-5,63){\rotatebox{90}{Accordion}}
% 	\put(-5,20){\rotatebox{90}{Dog}}
% 	\end{overpic}
% 	\vspace{8pt}
% 	\caption{
% 		\textbf{Visualization of the locations of embeddings in our pure co-representation and the channels of feature $\mathmat{A}^t$ transformed by our pure co-representation.}
% 		The first column are the inputs and their ground truth.
% 		In the first and third row, each of the red point ``\textcolor{red}{$\bullet$}" in one picture represents one of the top-k spatial positions whose corresponding representation constitutes the pure co-representation.
%         %
%         Each heatmap represents a channel of the converted feature (see \secref{sec:pcs}). 
%         It is a correlation map calculated from the deep representation of the target image and the representation noted in the above red point.
% 	}
% 	\vspace{-15pt}
% 	\label{fig:red_hot}
% \end{figure}

\subsection{Pure Co-Representation Search (PCS)}
\label{sec:pcs}

%To introduce PCS clearly, we 
Assume that we have obtained the co-representation proxy 
$\mathvec{p}^t\in\mathbb{R}^D$ of $t$-th iteration from RPP 
(see \secref{sec:rpp}).
Here, co-representation proxy $\mathvec{p}^t$ is a semantic embedding 
dominated by co-salient objects.
\revise{Then, the group of deep representations 
$\mathset{F}=\{\mathmat{F}_n\in\mathbb{R}^{D\times H\times W}\}_{n=1}^N$
contain $NHW$ pixel embeddings
$\{\mathset{F}^{(l)}\in\mathbb{R}^D\}_{l=1}^{NHW}$.
From the pixel embeddings,
PCS aims to find top-k embeddings $\{\mathvec{c}_k^t\in\mathbb{R}^D\}_{k=1}^K$, 
which are most correlated with the co-saliency dominated 
co-representation proxy $\mathvec{p}^t$.
Concretely, we calculate the correlation score between each pixel embedding 
$\mathset{F}^{(l)}\in\mathbb{R}^D$ and $\mathvec{p}^t$ by 
\begin{equation}\label{eq:score}
\textit{Score}^{t(l)} = \mathvec{p}^{t}\mathset{F}^{(l)\top}.
\end{equation}
After sorting the scores in descending order,
we record the spatial positions $\textit{Index}^{t}$ of the top-k embeddings 
with the highest correlation scores by
\begin{equation}
  \textit{Index}^{t} = \arg\mathop{\text{topk}}_l
	{\left(\textit{Score}^{t(l)}\right)} \in\mathbb{R}^K.
\label{eq:index}
\end{equation}
According to the positions $\textit{Index}^{t}$,
we gather the corresponding top-k embeddings from the $\mathset{F}$ 
as our co-representation,
\begin{equation}\label{eq:gather}
  \mathmat{C}^t = \text{gather}\left(\mathset{F}, \textit{Index}^t\right)
	\in\mathbb{R}^{K\times D}.
\end{equation}}

Once obtaining co-representation $\mathmat{C}^t$, 
we preliminarily filter noise by co-representation proxy $\mathvec{p}^t$ 
and then transform each feature $\mathmat{F}_n$ in $\mathset{F}$
to a set of correlation maps $\mathmat{A}_n\in\mathbb{R}^{K\times H\times W}$ 
by co-representation $\mathmat{C}^t$.
Specifically,
\begin{equation}
  \mathmat{\tilde{A}}_n^t = \mathmat{C}^{t}\left(
		\left(\mathvec{p}^t\mathmat{\tilde{F}}_n\right)
		\odot\mathmat{\tilde{F}}_n
	\right) \in\mathbb{R}^{K\times HW},
\label{eq:guide}
\end{equation}
where $\mathmat{\tilde{F}}_n\in\mathbb{R}^{D\times HW}$ is reshaped 
from $\mathmat{F}_n\in\mathbb{R}^{D\times H\times W}$.
We reshape the $\mathmat{\tilde{A}}^t_n\in\mathbb{R}^{K\times HW}$ 
to $\mathmat{A}^t_n\in\mathbb{R}^{K \times H\times W}$.
then we get $\mathset{A}^t = \{\mathmat{A}^t_n\}^{N}$ for the group of images.
Finally, we decode $\mathset{A}^t$ to predict the co-saliency maps 
$\mathset{M}^t$.

Further, we explain the transformed feature $\mathset{A}$ in detail.
Each $\mathmat{A}^t_n\in\mathbb{R}^{K\times H\times W}$ in $\mathset{A}^t$ 
can be regards as $K$ correlation maps calculated with $K$ embeddings 
$\{\mathvec{c}_k^t\in\mathbb{R}^{ D}\}_{k=1}^{K}$ 
of our co-representation $\mathmat{C}^t$.
In \figref{fig:red_hot},
we visualize some correlation maps and the locations (noted by ``red point" ) 
of their corresponding embeddings.
This transformed feature $\mathmat{A}^t_n$ brings three advantages 
according to the three observations below.

\noindent
1) \textbf{Sparse locations falls in the correct regions of co-salient objects.}
It means our co-representation $\mathmat{C}^t$ compose of the embeddings 
$\{\mathvec{c}_k^t\}_{k=1}^{K}$ extracted from the semantic embeddings 
belonging to co-salient objects.
That is why our co-representation is purer than current methods 
which aggregate information from  all locations;

\noindent
2) \textbf{Different embedding vectors $\mathvec{c}_k^t$ focus on different 
regions of co-salient objects.}
As the first example (person) in \figref{fig:red_hot},
the vectors extracted from the people's heads, chest, and legs give more 
activation to the corresponding regions of the target people.
Similarly, in the second group (person and dolphin),
the vectors of dolphins and the vectors of people, 
respectively activate the regions of the two kinds of objects.
Overall, the vectors $\mathvec{c}_k^t$ learn the information of the 
co-salient objects. 
For each vector, 
the information it learns is related to the position where it is exacted from.
This diversity helps CoRP predict a more comprehensive target object regions;
 
\noindent
3) \textbf{Semantic feature $\mathmat{F}_n$ converts to a concatenated 
correlation maps $\mathmat{A}^t_n$}, 
which is generated only from the correlation of intra group representation.
This conversion makes our CoRP focus on discovering intra group connections, 
rather than fitting the semantic categories of the training set.
All these advantages make our model obtain better predictions. 
We will further corroborate this in the ablation studies.

\begin{figure}[t]
	\centering
	\begin{overpic}[width=\columnwidth]{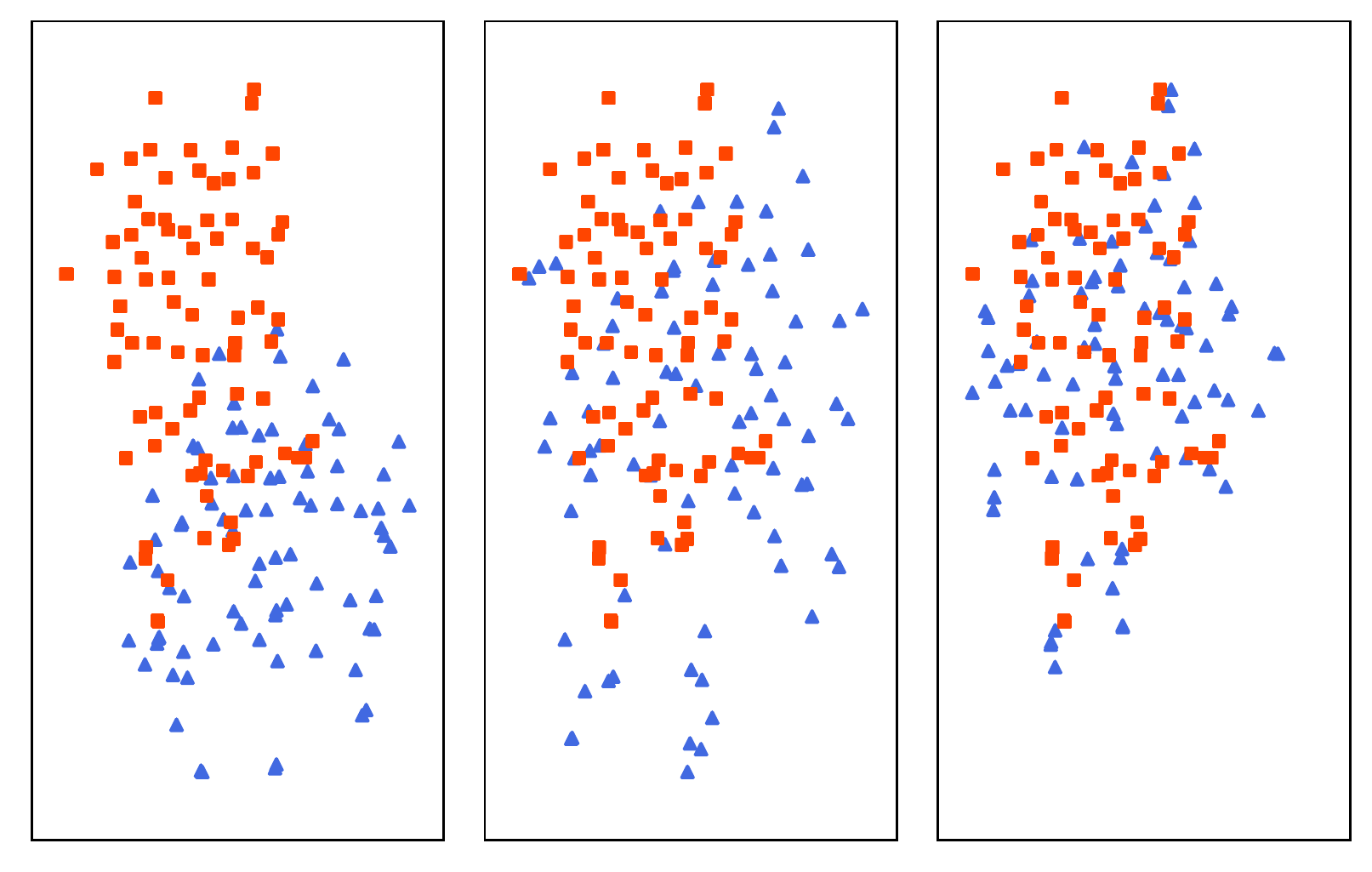}
	  \put(8.8, -2.5){Iteration 1}
	  \put(42.3, -2.5){Iteration 2}
	  \put(76, -2.5){Iteration 3}
	\end{overpic}
	\caption{\textbf{Distribution of the obtained co-representation proxies 
	  and corresponding ground-truth ones in different iterations.}
		We visualize all 80 groups in \coca by $t$-SNE~\cite{van2008tsne} algorithm.
		Here, the blue triangle ``\revise{$\blacktriangle$}" means the 
		co-representation proxies masked-averaged by co-saliency maps,
		and the orange square ``\textcolor{orange}{$\blacksquare$}" denotes 
		the pure co-representation proxies masked-averaged by ground-truth maps.
    The result shows that the iteration process makes the co-representation 
		proxies approach the ground-truth ones gradually.
	}\label{fig:iters}
  \vspace{-10pt}
\end{figure}

\begin{figure}[ht]
	\centering
	\begin{overpic}[width=\columnwidth]{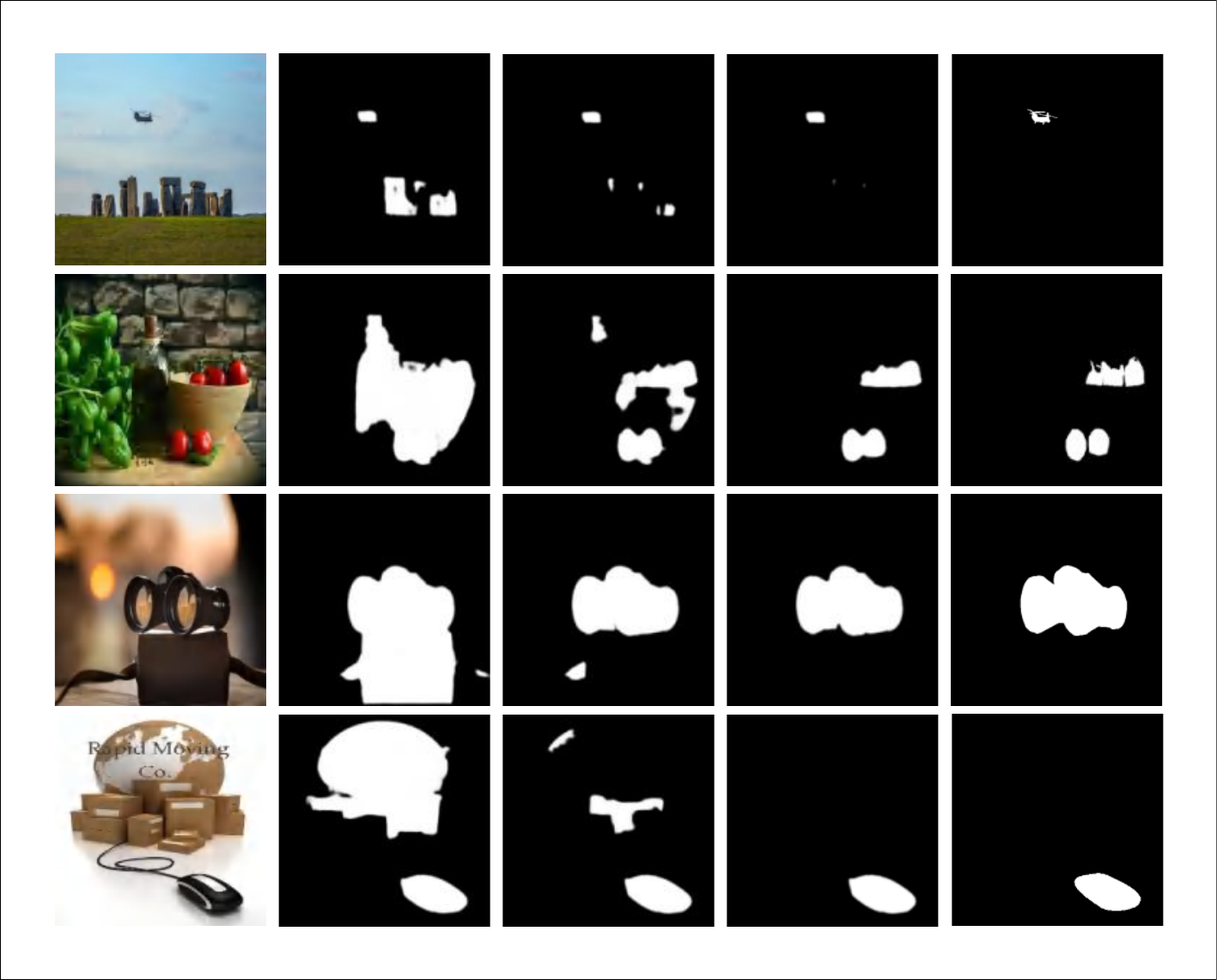}
		\put(5,-4){Input}
		\put(21,-4){Iteration1}
		\put(41,-4){Iteration2}
		\put(62,-4){Iteration3}
		\put(87,-4){GT}
	\end{overpic} \\ \vspace{5pt}
	\caption{\textbf{Predicitons in different iterations.}
		We can see that the predictions are gradually approaching the 
		ground-truth labels.
		It is thanks to the gradually purified co-representation in the iteration.
		Note that the predictions with less foreground noise are conducive 
		to mine purer co-representation. 
	}\vspace{-10pt}
	\label{fig:recurrent}
\end{figure}
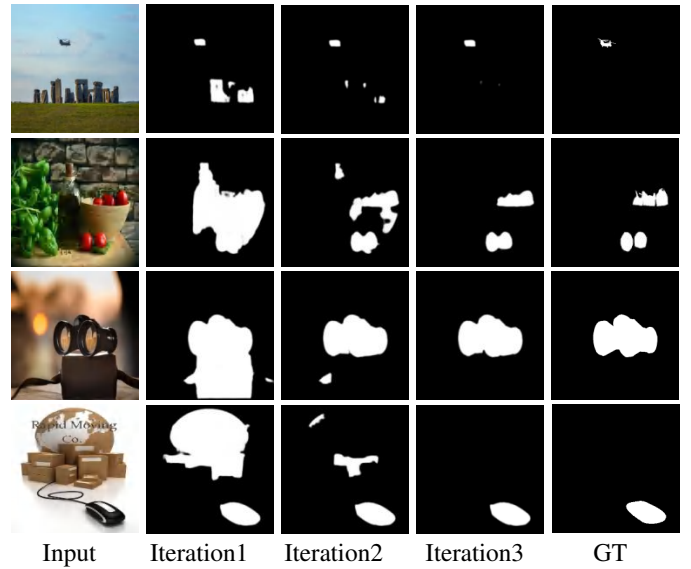

\subsection{Recurrent Proxy Purification (RPP)}
\label{sec:rpp}

RPP calculates co-representation proxy $\mathvec{p}^{t}\in\mathbb{R}^D$ 
based on the predicted co-saliency map 
$\mathset{M}^{t-1}=\{\mathmat{M}^{t-1}_n\}_{n=1}^N$ in the last iteration.
$\mathset{M}^0$ is initialized by our SOD head which shares backbone 
features with our Co-SOD network.
The details of the SOD head will be explained in \secref{Details}.
As co-salient object(s) recurs in every sample, 
the semantic representation masked by $\mathset{M}^{t-1}$ are domainated 
by the embeddings of co-salient object(s).
In this way,
we directly average the spatial locations highlighted by $\mathset{M}^{t-1}$ 
of $\mathset{F}$ as the co-representation proxy $\mathvec{p}^{t}$,
\begin{equation}\label{eq:gap}
  \mathvec{p}^t = \sum_{n=1}^N
	\frac{\mathtt{GAP}(\mathmat{M}^{t-1}_n\odot\mathmat{F}_n)}{N} \in\mathbb{R}^D,
\end{equation}
where $\odot$ denotes element-wise production, 
and the global average pooling (GAP) can be formalized as
$\mathtt{GAP}(\mathmat{F})=\frac{1}{HW}\sum_i\sum_j \mathmat{F}$, 
in which $i=1,...,W$ and $j=1,...,H$.
Again, $\mathvec{p}^{t}$ is normalized with Euclidean distance.

Note that $\mathvec{p}^{t}$ helps PCS predict more noise-free $\mathset{M}^{t}$ 
compared with $\mathset{M}^{t-1}$.
In turn, the $\mathset{M}^{t}$ makes RPP generate $\mathvec{p}^{t+1}$ 
with less noise compared with $\mathvec{p}^{t}$.
Working with pure co-representation search (PCS), 
recurrent proxy purification (RPP) makes our CoRP iterative like
\begin{equation}
\left\{
\begin{array}{rl}
\mathvec{p}^{t} &= \text{RPP}\left(\mathset{M}^{t-1}, \mathset{F}\right) \\
\mathset{M}^{t} &= \text{PCS}\left(\mathvec{p}^{t}, \mathset{F}\right) 
\end{array}
\right., t=\{1,2,3,...,T\}.
\end{equation}
Besides, we just utilize the encoder of our model for one time and 
the encoder does not paticipate in the following iteration process.
In other words, our RPP makes PCS and the decoder works repeatly without encoder.
In this case, the iteration process does not increase much computational burden.
We details the iterative process in \algref{alg:corp}.

\begin{algorithm}[t]
  \SetAlgoLined
  \SetKwBlock{RPP}{RPP}{end}
  \SetKwBlock{PCS}{PCS}{end}
  \textbf{Input: } A group of images $\mathset{I}=\{\mathmat{I}_n\}_{n=1}^N$

  \textbf{Output: } Co-saliency maps 
	  $\mathset{M}^T=\{\mathmat{M}^T_n\}_{n=1}^N$

  \textbf{Initialize: } Extract deep representation 
	  $\mathset{F}=\{\mathmat{F}_n\}_{n=1}^N$; 
		initilize $\mathset{M}^0$ with our backbone-shared SOD head

  \For{$t\leftarrow 1$ \KwTo $T$}{
  \RPP{
    Generate co-representation proxy $\mathvec{p}^{t}$ based on $\mathset{F}$ 
		and $\mathset{M}^{t-1}$~(\eqref{eq:gap})
  }
  \PCS{
    Obtain $\textit{Score}^{t}$ by scoring each spatial location in 
		$\mathset{F}$ with $\mathvec{p}^{t}$~(\eqref{eq:score})

    Search top-k purer embeddings as co-representation $\mathmat{C}^{t}$ 
		from $\mathset{F}$ based on $\textit{Score}^{t}$~(\eqref{eq:index}, 
		\ref{eq:gather})

    Transform each $\mathmat{F}_n$ to a set of correlation maps 
		$\mathmat{A}^t_n$ with co-representation 
		$\mathmat{C}^{t}$~(\eqref{eq:guide}), 
		then obtain $\mathset{A}^t = \{\mathmat{A}^t_n\}^{N}$

    Predict $\mathset{M}^t=\{\mathmat{M}^t_n\}_{n=1}^N$ 
		by decoding $\mathset{A}^t$
  }
  }
  \caption{Co-Salient Object Detection with Co-Representation Purification}
  \label{alg:corp}
\end{algorithm}

To better understand what happened in the iteration,
we analyze the changes of three important factors.

\noindent
1) \textbf{Co-representation proxy} gradually approaches that obtained 
by ground-truth annotation.
In \figref{fig:iters},
we visualize the co-representation proxies of all 80 groups in \coca by $t$-SNE.
In each iteration,
we denote the proxies masked-averaged by co-saliency maps as ``blue triangle",
and these masked-averaged by ground-truth annotation as ``orange square".
With the co-saliency maps are gradually predicted accuracy, 
the distance between the two kinds of  proxies gradually shrinks.
As the orange nodes denote noise-free proxies completely generated 
by embeddings belonging to co-saliency objects, 
the reduction of the distance means our RPP gradually reduces the
distracting information in our proxies. 

\noindent
2) \textbf{Co-representation} consists of more and more repesenations 
from co-salient regions.
In \tabref{tab:percentage},
we count the proportion of the space-wise position of the embeddings falling 
on the co-salient object.
On the three benchmark datasets, 
the average proportions grow after each iteration.
After three-times of iteration, more than two-thirds of embeddings 
in the co-representations are located at co-salient objects.
The growth of proportions means the co-representations contain more co-saliency 
information and less distracting noise.
Due to the reduction of distracting information in the proxy, 
the co-representation is gradually purified with the iterative process.

\noindent
3) \textbf{Co-saliency maps} gradually eliminate the background and 
irrelevant foreground objects.
We show the predicted results in \figref{fig:recurrent}.
In the first iteration, background interference is suppressed, 
but it still contains much foreground noise, 
which fades away during further interations guided by purer co-representation.

\begin{table}[t]
	\tablestyle{4.1mm}{1.2}
  \caption{\textbf{Proportion of the sparse locations for extracting 
	  co-representation falling on the co-salient objects.} 
    With the iterative process, 
    co-representation extracts more from the corresponding feature positions 
		of co-salient objects;
    thus, our co-representation becomes purer and purer.
  }
  \begin{tabular}{l|c|c|c} \hline                    
    & CoCA~\cite{zhang2020gicd}
    & CoSal2015~\cite{zhang2016CoSal} & CoSOD3k~\cite{fan2020taking} \\ \hline
		$T$ = 1 & 60.3\% & 93.2\% & 82.4\% \\ \hline
		$T$ = 2 & 65.3\% & 94.6\% & 84.8\% \\ \hline
		$T$ = 3 & 66.8\% & 94.9\% & 85.1\% \\ \hline
		$T$ = 4 & 67.2\% & 95.0\% & 85.2\% \\ \hline
		$T$ = 5 & 67.3\% & 95.0\% & 85.3\% \\ \hline
		$T$ = 6 & 67.3\% & 95.0\% & 85.3\% \\ \hline
   \end{tabular}
   \vspace{-5pt}
   \label{tab:percentage}
\end{table}

\section{Experiment}
\subsection{Implementation Details}\label{Details}

\myPara{Model details}
We employ pretrained VGG-16 \cite{simonyan2015vgg} as our backbone and construct an encoder-decoder architecture.
In CoRP, the initial saliency map $\mathset{M}^0$ is predicted by our backbone-shared SOD head.
After acquiring the first co-representation proxy $\mathvec{p}^{1}$ with $\mathset{M}^0$, we employ our PCS on the last four outputs of the encoder.
PCS aims at localizing co-saliency objects and produces four co-saliency features. 
Our decoder, which is the same as that of ICNet~\cite{Jin2020ICNet} decodes the four co-saliency features and remaining shallow features to predict co-saliency maps $\mathset{M}^1$.
Our RPP feeds back $\mathset{M}^1$ to produce a new co-representation proxy and repeat the process just mentioned to predict $\mathset{M}^2$.
The iteration process produces $\{\mathset{M}^1, \mathset{M}^2, \mathset{M}^3,... ,\mathset{M}^T\}$ and we set $T=3$ in our experiments. 
The reason will be explained in \secref{ablations}.
% Specifically, we denote the outputs of VGG-16 encoder with an adding convolution layer as $\{\mathset{F}_1, \mathset{F}_2,\mathset{F}_3,\mathset{F}_4,\mathset{F}_5,\mathset{F}_6\}$. 
% In CoRP, the initial saliency map $\mathset{M}^0$ which is predicted by our backbone weights shared SOD head.
% After acquiring first co-representation proxy $\mathvec{p}^{1}$ with $\mathset{M}^0$, we apply our PCS on
% $\{\mathset{F}_3,\mathset{F}_4,\mathset{F}_5,\mathset{F}_6\}$ and acquire four co-saliency features.
% To explore background consistency, we also utitlize inversed saliency maps $1-\mathset{M}^0$ to acquire background four co-saliency features.
% In decoding stage, after concatenating the coresponding foreground and background features, the four multi-scale concatenated features are merged to generate co-saliency feature $\mathset{F}_M$, we directly decode $\mathset{F}_M$ to acquire primary prediction $\mathset{M}_p$.
% To acquire more accurate prediction, we multiply $\{\mathset{F}_1, \mathset{F}_2\}$ with $\{\mathset{M}_p$, $\mathset{M}^0\}$ and concatenate them with the merged co-saliency feature $\mathset{F}_M$.
% This feature is decoded to get refined prediction $\mathset{M}^1$, our RPP feed back $\mathset{M}^1$ into PCS and repeat the process just metioned to predict $\mathset{M}^2$.
% The iteration process produces $\{\mathset{M}_1, \mathset{M}_2, \mathset{M}_3,... ,\mathset{M}_T\}$ and we choose $\mathset{M}_3$ as our final prediction.

The SOD head, which produces initial saliency maps $\mathset{M}^0$ shares backbone weights with our CoSOD network.
The decoder of the SOD head is designed independently and directly merges the outputs of the encoder and predict saliency maps.
The SOD head adds 2.9 MB of model parameters to our CoRP.

\begin{table*}[t!]
	\tablestyle{0.39mm}{1.3}
	\caption{
		\textbf{Quantitative comparisons} of \revise{mean
		absolute error ($MAE$)},
		maximum F-measure~\cite{borji2015SalObjBenchmark} ($F_{\max}$),
		S-measure~\cite{fan2017structure} ($S_{\alpha}$),
		and mean E-measure~\cite{Fan2018Em} ($E_{\xi}$) by our CoRP and other methods on the
	\textit{CoCA}~\cite{zhang2020gicd}, \textit{CoSOD3k}~\cite{fan2020taking}, and \textit{CoSal2015}~\cite{zhang2016CoSal} datasets. 
	 \revise{DUTS-Class \cite{zhang2020gicd}, COCO-9k \cite{wei2019group-wise}, COCO-SEG \cite{zha2020robust},
	 and  MSRA-B \cite{liu2010learning} are wildly used training datasets in Co-SOD and we denote them as Train-1, 2, 3, and 4, respectively.}
		``$\uparrow$'' means that the higher the numerical value, the better the model performance.
		The numerical value with an underline refers to the second best result under VGG16 backbone.
		Specifically, our CoRP shows the results in the third iteration ($T$ = 3).
		We provide the performances of our model with three backbones: VGG16\cite{simonyan2015vgg}, ResNet50 \cite{he2016deep} and PVTv1-medium \cite{wang2021pyramid}.
		\revise{CoRP$^\text{NAS}$ represents that we find the optimal hyper parameter K in our PCS by Network Architecture Search \cite{guo2020single}. }
	}
	\begin{tabular}{lr|cc|ccccccccc|ccccc}
		\hline 
		&  & GateNet & GCPA         &  CBCD   &   CSMG  & GCAGC  &  GICD   & ICNet & CoEG & \revise{DeepACG} & GCoNet & \revise{CADC} & \revise{CoRP} & CoRP & \revise{CoRP}$^\text{\revise{NAS}}$ & CoRP & CoRP\\
		[-0.2cm]
		& & \scriptsize ECCV20 & \scriptsize AAAI20    & \scriptsize TIP13   &  \scriptsize CVPR19  & \scriptsize CVPR20  & \scriptsize ECCV20   & \scriptsize  NeurIPS20 & \scriptsize TPAMI21 & \scriptsize \revise{CVPR21} & \scriptsize CVPR21  & \scriptsize \revise{ICCV21} & \revise{\scriptsize 2021} & \scriptsize 2021 & \revise{\scriptsize 2021} & \scriptsize 2021 & \scriptsize 2021  \\
		[-0.2cm]
		&  & \cite{Zhao2020GateNet} & \cite{Chen_Xu_Cong_Huang_2020GCPANet}   & \cite{Fu2013ClusterCo} &  \cite{zhang2019csmg} &  \cite{Zhang2020GCAGC}  & \cite{zhang2020gicd}  & \cite{Jin2020ICNet} & \cite{deng2021re} & \revise{\cite{zhang2021deepacg}} & \cite{fan2021GCoNet} & \revise{\cite{zhang2021summarize}} & \revise{VGG16} & VGG16 & \revise{VGG16} & Res-50 & PVT \\
		& \revise{Training Set}  & \revise{-} & \revise{-}   & \revise{-} &  \revise{\scriptsize{Train-4}} & \revise{\scriptsize{Train-3}}   & \revise{\scriptsize{Train-1}} & \revise{\scriptsize{Train-1,2}} & \revise{\scriptsize{Train-1}} & \revise{\scriptsize{Train-3}} & \revise{\scriptsize{Train-1}} & \revise{\scriptsize{Train-1,2}} & \revise{\scriptsize{Train-2}} & \revise{\scriptsize{Train-1,2}} & \revise{\scriptsize{Train-1,2}} & \revise{\scriptsize{Train-1,2}} & \revise{\scriptsize{Train-1,2}} \\
%		\rowcolor{gray!10}
		\hline
%		\cellcolor{white}
		& \revise{MAE$\downarrow$}  & \revise{0.097} & \revise{0.082}     & \revise{0.233}  &  \revise{0.130} & \revise{0.085} & \revise{0.071}   & \revise{\underline{0.058}} & \revise{0.078} & \revise{0.064} & \revise{0.068} & \revise{0.064} & \revise{0.060} & \revise{0.049} & \revise{\textbf{0.049}} & \revise{0.046} & \revise{0.044}\\
		& $F_{\max}\uparrow$  & 0.772 & 0.830   & 0.547  & 0.787 & 0.832 &  0.844   & 0.859 & 0.836 & \revise{0.842} & 0.847 & \revise{\underline{0.862}} & \revise{0.864}  & 0.885 & \revise{\textbf{0.888}} & \revise{0.893} & \revise{0.895}\\
%		\rowcolor{gray!10} \cellcolor{white}
		& $S_{\alpha}\uparrow$  & 0.811 & 0.850    & 0.550  &  0.776  & 0.823 &   0.844   & 0.855 & 0.838  & \revise{0.854} & 0.845 & \revise{\underline{0.866}} & \revise{0.859} & 0.875 & \revise{\textbf{0.877}} & \revise{0.879} & \revise{0.884}\\
		\multirow{-4}{*}{\begin{sideways}CoSal2015\end{sideways}}
		& $E_{\xi}\uparrow$  & 0.820 & 0.864      & 0.516  &  0.763   & 0.814 &   0.883   & \underline{0.896} & 0.868 & \revise{-}  & 0.884 & \revise{-} & \revise{0.896} & 0.913 & \revise{\textbf{0.915}} & \revise{0.919} & \revise{0.920}\\
%		\rowcolor{gray!10} 
		\hline
%		\cellcolor{white}
		& \revise{MAE$\downarrow$} & \revise{0.173} & \revise{0.188}      & \revise{0.180}  &  \revise{0.114}  & \revise{0.111}   & \revise{0.126}  & \revise{0.148} & \revise{0.106} & \textbf{\revise{0.102}} & \revise{\underline{0.105}} & \revise{0.132} & \revise{0.101} & \revise{0.121} & \revise{0.110} & \revise{0.104} & \revise{0.093}\\ 
 		& $F_{\max}\uparrow$  & 0.398 & 0.435   & 0.313  & 0.499 & 0.517 &  0.513   & 0.514 & 0.493 & \revise{\underline{0.552}} & 0.544 & \revise{0.548} & \revise{0.564} & 0.551 & \revise{\textbf{0.575}} & 0.607 & 0.619 \\
%		\rowcolor{gray!10} \cellcolor{white}
		& $S_{\alpha}\uparrow$  & 0.600 & 0.612    & 0.523  &  0.627  & 0.666 & 0.658 & 0.657 & 0.612 & \revise{\underline{0.688}} & 0.673  & \revise{0.681} & \revise{0.699} & 0.686 & \revise{\textbf{0.703}} & 0.719 & 0.732 \\
		\multirow{-4}{*}{\begin{sideways}CoCA\end{sideways}}
		& $E_{\xi}\uparrow$    & 0.609 & 0.612      & 0.535  &  0.606  & 0.668 & 0.701 & 0.686 & 0.679 & \revise{-} & \underline{0.739} & \revise{-} & \revise{0.750}   & 0.715 & \revise{\textbf{0.741}} & 0.745 & 0.773\\
		\hline
%		\cellcolor{white}
		& \revise{MAE$\downarrow$}   & \revise{0.112} & \revise{0.104}    & \revise{0.228} & \revise{0.157} & \revise{0.100} & \revise{0.079} & \revise{0.097} & \revise{0.084}  & \revise{0.089} & \revise{\textbf{0.071}} & \revise{0.076} & \revise{0.067}  & \revise{0.075} & \revise{\underline{0.072}} & \revise{0.057} & \revise{0.057}\\
		& $F_{\max}\uparrow$   & 0.697 & 0.746   & 0.468 & 0.730 & \underline{0.779} & 0.770 & 0.766 & 0.758 & \revise{0.756} & 0.777 & \revise{0.759} & \revise{0.794} & 0.798 & \revise{\textbf{0.801}} & 0.828 & 0.835 \\
%		\rowcolor{gray!10} \cellcolor{white}
		& $S_{\alpha}\uparrow$   & 0.763 & 0.795    & 0.529 & 0.727 & 0.798 & 0.797 & 0.798 & 0.778  & \revise{0.792} & \underline{0.802}  & \revise{0.801} & \revise{0.820}  & 0.820 & \revise{\textbf{0.825}} & 0.842 & 0.850 \\
		\multirow{-4}{*}{\begin{sideways}CoSOD3k\end{sideways}}
		& $E_{\xi}\uparrow$   & 0.772 & 0.813       & 0.509 & 0.675 & 0.791 &    0.845 & 0.843 & 0.817 & \revise{-} & \underline{0.857}  & \revise{-}  & \revise{0.864} & 0.862 & \revise{\textbf{0.866}} & 0.887 & 0.891 \\
		\hline
	\end{tabular}
	\vspace{6pt}
	
	\label{tab:Results}
\end{table*}

\myPara{Training details} 
Following \cite{Jin2020ICNet}, the datasets we use for training our 
co-saliency model (CoRP) are DUTS dataset~\cite{wang2017supervision} 
and a subset of COCO dataset~\cite{Lin2014COCO}, containing 9213 images. 
The Co-SOD network is trained with the subset of COCO~\cite{Lin2014COCO} 
while the backbone shared SOD head is trained together with DUTS dataset
\cite{wang2017supervision}.
The Adam optimizer is used with an initial learning rate of 
$1\times 10^{-5}$ and a weight decay of $1\times 10^{-4}$.
We train our model for 70 epochs in total.
It takes about 4.5 hours on an Nvidia Titan X.
In each training iteration, we randomly select 10 samples as a batch from 
an image group of COCO~\cite{Lin2014COCO} subset to train the Co-SOD network 
and 8 samples from DUTS~\cite{wang2017supervision} to train SOD head.
At testing stage, each batch is made up of all images within a Co-SOD group, 
and images from SOD datasets are not required.
At both training and test stages, the images are resized into $224 \times 224$, 
and we augment the data by randomly flip images horizontally 
at the training stage.
Finally, the predictions are resized back for evaluation.
Our CoRP is implemented in PyTorch \cite{pytorch2019paszke} and 
Jittor \cite{hu2020jittor}.
The PyTorch version runs at 45.3 FPS with total parameters size of 80.1 MB 
on an Nvidia Titan X GPU when the number of iterations is set to three ($T$=3).

At training stage, we use the ground truth as mask to generate a 
noise-free co-representation proxy.
With a completely noise-free proxy, our PCS locate co-saliency objects 
accurately and this training strategy makes our Co-SOD network fully 
rely on PCS to locate objects.
In this way, our RPP is not employed at training stage.  
At testing stage, the first co-representation proxy is initialized by 
saliency maps produced by our SOD head.
At the same time, RPP iteratively feeds predicted co-saliency maps into PCS, 
which provides more accurate objects localization and finally generates more 
accurate co-saliency maps.
The iteration times of RPP can be set freely at testing stage, 
and we will discuss it in \secref{ablations}.

\begin{figure*}[t!]
	\centering
	\begin{overpic}[width=\textwidth]{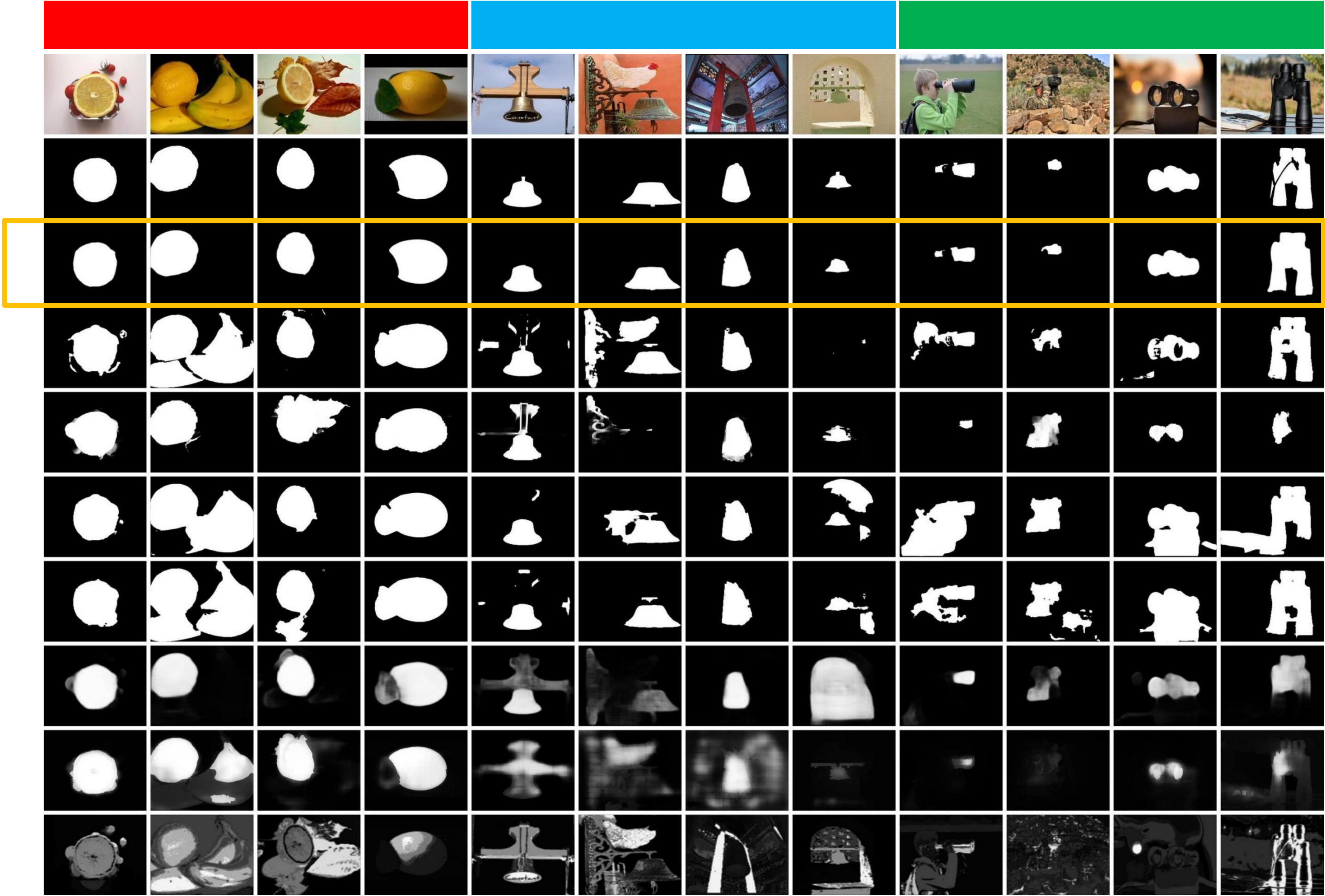}
	\put(14.2,65){\color{white}\textbf{CoSal2015}}
	\put(47.5,65){\color{white}\textbf{CoSOD3k }}
	\put(81.1,65){\color{white}\textbf{CoCA }}
	\put(1.7,58.7){\scriptsize{\rotatebox{90}{Input}}}
	\put(1.7,52.7){\scriptsize{\rotatebox{90}{GT}}}
	\put(1.7,46.0){\scriptsize{\rotatebox{90}{CoRP}}}
	\put(1.7,39.0){\scriptsize{\rotatebox{90}{GCoNet}}}
	\put(1.7,33.0){\scriptsize{\rotatebox{90}{CoEG}}}
	\put(1.7,26.9){\scriptsize{\rotatebox{90}{ICNet}}}
	\put(1.7,20.5){\scriptsize{\rotatebox{90}{GICD}}}
	\put(1.7,13.3){\scriptsize{\rotatebox{90}{GCAGC}}}
	\put(1.7,7.5){\scriptsize{\rotatebox{90}{CSMG}}}
	\put(1.7,1){\scriptsize{\rotatebox{90}{CBCD}}}
	\end{overpic} \\
	\vspace{-5pt}
	\caption{
		\textbf{Visual comparison} of our CoRP with six SOTA methods. 
		We demonstrate the predictions from three categories (lemon, chime, and binoculars) belonging to the
		three benchmark dataset (\textit{CoSal2015}~\cite{zhang2016CoSal},   \textit{CoSOD3k}~\cite{fan2020taking}, and \textit{CoCA}~\cite{zhang2020gicd}).
		We highlight the results of CoRP with an orange frame.
	}
	\label{fig:compare}
\end{figure*}
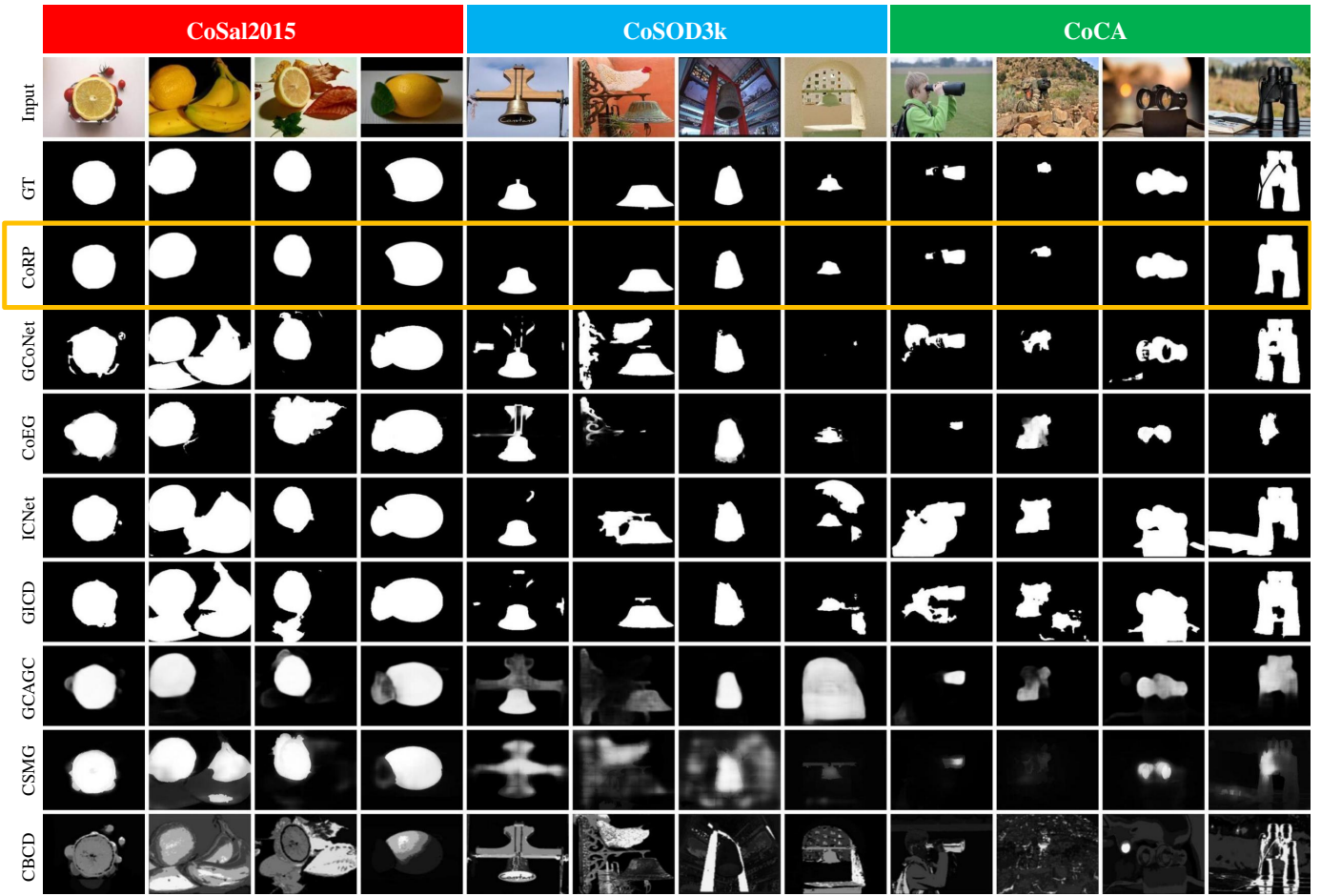

\begin{table*}[t]
	\tablestyle{1.9mm}{1.3}
	\caption{
		\textbf{Ablation study} of the proposed CoRP on the \textit{CoCA} and \textit{CoSOD3k} datasets.
		``SOD'' refers to our SOD head which share backbone weights with our Co-SOD network.
		``PCS'' and ``RPP'' are the proposed pure co-representation search and recurrent proxy purification.
        ``baseline'' refers to our Co-SOD encoder-decoder architecture without ``SOD'', ``PCS'' and ``RPP''.
		``baseline + SOD + proxy" denotes using directly the proxy as co-representation.
	}
	\begin{tabular}{c|l|cccc|cccc|cccc}
		\hline 
		%[-0.1cm]
		%[-0.3ex]
		\multirow{2}{*}{ID} & \multirow{2}{*}{Combination} & \multicolumn{4}{c|}{CoCA~\cite{zhang2020gicd}} & \multicolumn{4}{c|}{CoSal2015~\cite{zhang2016CoSal}}  & \multicolumn{4}{c}{CoSOD3k~\cite{fan2020taking}}   \\
		        &   & $F_{\text{avg}}\uparrow$ & $F_{\max}\uparrow$ & $S_{\alpha}\uparrow$ & $E_{\xi}\uparrow$ &  $F_{\text{avg}}\uparrow$ & $F_{\max}\uparrow$ & $S_{\alpha}\uparrow$ & $E_{\xi}\uparrow$ 
		        &  $F_{\text{avg}}\uparrow$ & $F_{\max}\uparrow$ & $S_{\alpha}\uparrow$ & $E_{\xi}\uparrow$\\
		%[-0.1cm]
		\hline
%		\rowcolor{gray!10}
		1  & baseline 
		& 0.381 & 0.397 & 0.560 & 0.563 & 0.652 & 0.662 & 0.699 & 0.727  & 0.576 & 0.590 & 0.654 & 0.687  \\
		2  & baseline + SOD                         
		& 0.436 & 0.443 & 0.608 & 0.640 & 0.760 & 0.772 & 0.800 & 0.830  & 0.688 & 0.697 & 0.753 & 0.795  \\
	    3  & baseline + SOD + proxy            
	    & 0.381 & 0.399 & 0.551 & 0.531 & 0.798 & 0.825 & 0.836 & 0.866  & 0.675 & 0.697 & 0.744 & 0.766  \\
%		\rowcolor{gray!10}
		4  & baseline + SOD + PCS 
		& 0.474 & 0.488 & 0.640 & 0.658 & 0.860 & 0.874 & 0.869 & 0.906  & 0.753 & 0.764 & 0.800 & 0.839   \\ \hline
		CoRP  & baseline + SOD + PCS + RPP  
		& 0.541 & 0.551 & 0.686 & 0.715 & 0.872 & 0.885 & 0.875 & 0.913 & 0.788 & 0.798 & 0.820 & 0.862  \\
		\hline
	\end{tabular}
    \vspace{5pt}
	\label{tab:ABL}
\end{table*}

\myPara{Loss function} 
We take the IoU loss~\cite{lin2019agss} for training our CoRP. 
The IoU loss has been proved to be effective for Co-SOD task in~\cite{zhang2020gicd,Jin2020ICNet}.
% In order to detect the position of co-salient objects and separate them from the background and the other non co-salient foreground accurately, we supervise the prediction under IoU loss \cite{lin2019agss} while training our Co-SOD model (CoRP). 
Its specific formula is as follows:
\begin{equation}
\text{L}(\mathset{M}, \mathset{G}) \!=\! 1 \!-\! \sum_{n=1}^N \frac{\sum_{w=1}^W \sum_{h=1}^H \text{min}\{\mathmat{M}^{w,h}_n,\mathmat{G}^{w,h}_n\}}{\sum_{w=1}^W \sum_{h=1}^H \text{max}\{\mathmat{M}^{w,h}_n,\mathmat{G}^{w,h}_n\}}.
\end{equation}
$\mathset{M} = \{\mathmat{M}_n\}_{n=1}^N$ and $\mathset{G} =  \{\mathmat{G}_n\}_{n=1}^N$ represents the co-saliency maps generated by our model and the ground truth respectively. 
$\mathmat{M}^{w,h}_n$ denotes a pixel position of the $n$-th co-saliency map $\mathmat{M}_n$ in a group of predictions, and so does $\mathmat{G}^{w,h}_n$ in the corresponding group of ground truth. 

To supervise our SOD head, we use $\mathset{S} = \{\mathmat{S}_m\}_{m=1}^M$ and $\mathset{T} =  \{\mathmat{T}_m\}_{m=1}^M$ to represent the predicted saliency maps and the corresponding ground truth.
We also employ IoU loss $\text{L}(\mathset{S}, \mathset{T})$ for the SOD head.
Our final loss function for training both Co-SOD network and SOD head simultaneously is as follows:
\begin{equation}
\text{L}_{all} = \alpha \text{L}(\mathset{M}, \mathset{G}) + \beta \text{L}(\mathset{S}, \mathset{T}),
\end{equation}
where we set $\alpha=0.8$ and $\beta=0.2$.

\subsection{Evaluation Datasets and Metrics}

\myPara{Datasets}
% We evaluate our CoRP on three challenging datasets:
% \textit{CoSal2015} \cite{zhang2016CoSal}, \textit{CoSOD3k} \cite{fan2020taking}, and \textit{CoCA} \cite{zhang2020gicd}.
% The three datasets are widely used in evaluating Co-SOD methods.
% It is worthy of mentioning that in \textit{CoCA} \cite{zhang2020gicd}, nearly every image contains extraneous salient objects except for co-salient objects.
% Meanwhile,
% the image groups in \textit{CoCA} \cite{zhang2020gicd} dataset are entirely unseen category for our CoRP because \textit{CoCA} does not have the same category as our training set.
% Thanks to these two key factors, 
% the results on \textit{CoCA} can better reflect the performance of each method on the class-agnostic Co-SOD task.
\revise{We evaluate our CoRP on three challenging datasets:
\textit{CoSal2015} \cite{zhang2016CoSal}, \textit{CoSOD3k} \cite{fan2020taking}, and \textit{CoCA} \cite{zhang2020gicd}.
The three datasets are widely used in evaluating Co-SOD methods.
\textit{Cosal2015} \cite{zhang2016CoSal} contains 2015 images of 50 categories, 
and each group has one or more challenging issues such as complex environments, object appearance variations, occlusion,
and background clutters. 
\textit{CoSOD3k} \cite{fan2020taking} is the largest CoSOD evaluation dataset so
far, 
which covers 13 super-classes and contains 160 groups and totally 3,316 images.
Each group includes diverse realistic scenes, and different object appearances, and covers the major challenges in CoSOD.
\textit{CoCA} \cite{zhang2020gicd} dataset consists of 80 categories with 1295
images
Compared with \textit{Cosal2015} and \textit{CoSOD3k},  
\textit{CoCA} has two special characteristics.
Firstly, every image contains extraneous salient objects except for co-salient objects.
Meanwhile,
the image groups in \textit{CoCA} \cite{zhang2020gicd} dataset are entirely unseen categories for our CoRP because \textit{CoCA} does not have the same category as our training set.
Thanks to these two key factors, 
the results on \textit{CoCA} can better reflect the performance of each method on the class-agnostic Co-SOD task.}

\myPara{Metrics} 
Following GICD~\cite{zhang2020gicd}, we report four widely used metrics, 
namely \revise{mean absolute error (MAE)}, 
maximum F-measure ($F_{\max}$)~\cite{borji2015SalObjBenchmark}, 
S-measure ($S_{\alpha}$)~\cite{fan2017structure}, 
and mean E-measure ($E_{\xi}$).
The evaluation code is available at 
\url{https://github.com/zzhanghub/eval-co-sod}.

\subsection{Comparison with Current Methods}
\revise{To illustrate the performance of our CoRP, we compare our method with nine SOTA Co-SOD methods, including CBCD~\cite{Fu2013ClusterCo}, CSMG~\cite{zhang2019csmg}, GCAGC~\cite{Zhang2020GCAGC}, GICD~\cite{zhang2020gicd}, CoEG-Net~\cite{deng2021re}, ICNet~\cite{Jin2020ICNet}, DeepACG~\cite{zhang2021deepacg}, GCoNet~\cite{fan2021GCoNet}, and CADC~\cite{zhang2021summarize}. 
Meanwhile, following \cite{zhang2020gicd, Jin2020ICNet, zhang2020CoADNet}, we also report the result of two SOTA SOD methods, namely GateNet~\cite{Zhao2020GateNet} and GCPA~\cite{Chen_Xu_Cong_Huang_2020GCPANet}, as baselines.}

\myPara{\revise{Quantitative evaluation.}}
\revise{
In \tabref{tab:Results}, we compare our method CoRP with other SOTA Co-SOD and SOD methods in quantitative results.
We can see that our CoRP achieves SOTA performances across the three benchmark datasets.
On \textit{CoCA}~\cite{zhang2020gicd} dataset, the performance of our method surpasses the others to a great degree in terms of max F-measure and S-measure.
Compared with the second-best methods on \textit{CoCA}~\cite{zhang2020gicd} dataset, 
our method CoRP lead them by 1.5\% in terms of S-measure and 2.3\% in terms of max F-measure.
As there are extraneous salient objects except for co-salient objects in each image of \textit{CoCA}, 
our outperformance shows that our method has a better capacity for localizing co-salient objects.
Meanwhile, \textit{CoCA}~\cite{zhang2020gicd} does not contain any image category which appears in our training dataset \textit{COCO}~\cite{Lin2014COCO}.
Our leading performance on this dataset demonstrates that our method is robust to the unseen class in the class-agnostic task.
Noticing that the performance of SOD methods, especially
GCPANet~\cite{Chen_Xu_Cong_Huang_2020GCPANet}, is comparable to or even better than some Co-SOD methods on \textit{CoSOD3k}~\cite{fan2020taking} and \textit{CoSal2015}~\cite{zhang2016CoSal} datasets because a large part of images in these two datasets contain only one salient object.
However, our method also surpasses the second-best Co-SOD methods on \textit{CoSal2015}~\cite{zhang2016CoSal} and  \textit{CoSOD3k}~\cite{fan2020taking} by 1.1\% and 2.3\% in terms of S-measure. 
Our performance on the three benchmark datasets fully demonstrates our method's effectiveness.
In addition,
our method does not outperform all the others in terms of MAE.
This may be due to the insufficient capacity of our method to segment object details.}

\myPara{Qualitative result.}
In \figref{fig:compare}, we compare the our method's predictions with other methods on the three benchmark datasets.
Our method successfully separates the lemons from other parts of the images, 
while some other methods fail to distinguish the bananas from the lemons.
Although the images of chime have complex background with great change, our method localizes the chimes and separates them from the background accurately.
Meanwhile, the hands on the binocular are tiny but our method can separate the binoculars from the hands.

\begin{figure*}[t!]      
	\centering
	\begin{overpic}[width=1\textwidth]{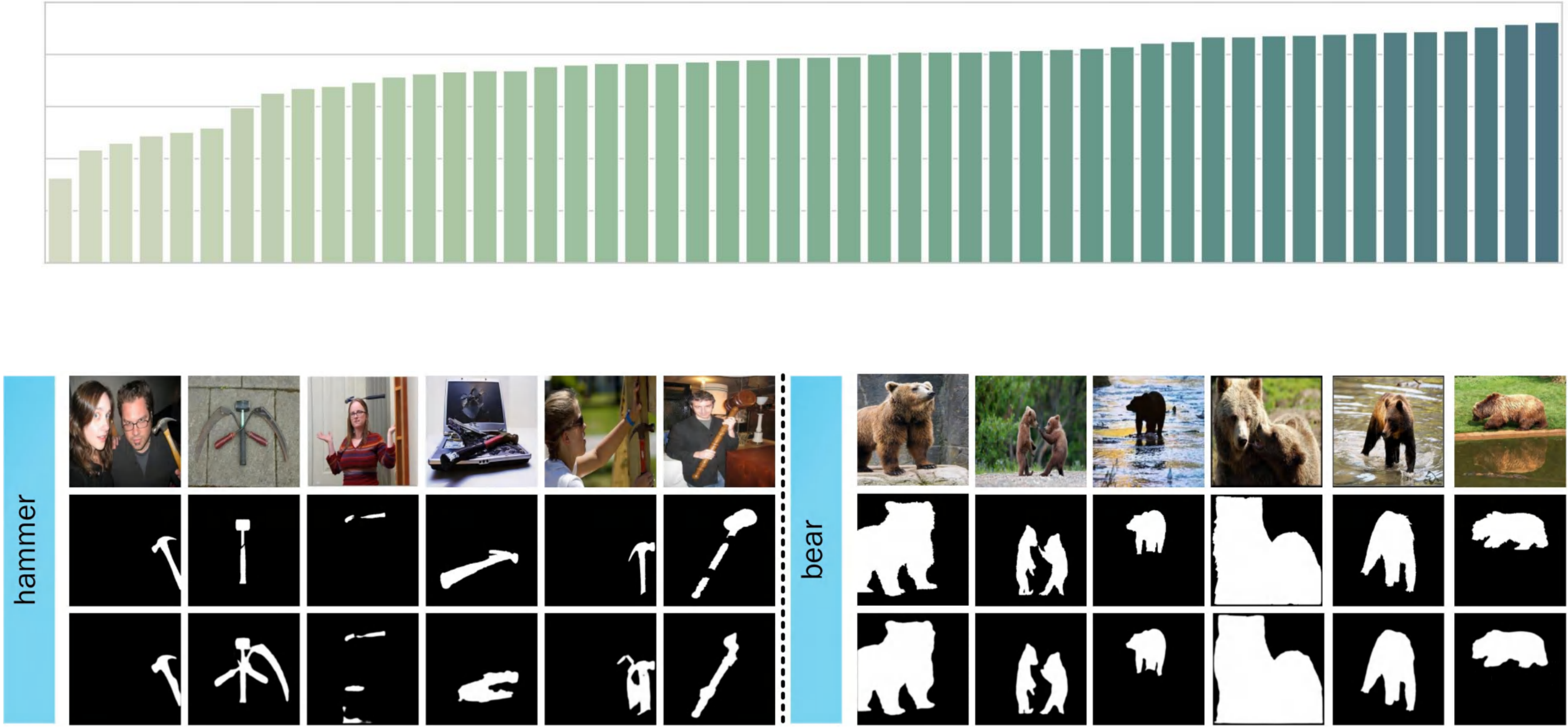}
	\put(3.2, 24.5){\scriptsize{\rotatebox{90}{hammer}}}
	\put(5.2, 23.7){\scriptsize{\rotatebox{90}{pineapple}}}
	\put(7.2, 23.1){\scriptsize{\rotatebox{90}{frenchhorn}}}
	\put(9.2, 26.0){\scriptsize{\rotatebox{90}{piano}}}
	\put(11.2, 24.5){\scriptsize{\rotatebox{90}{baseball}}}
	\put(13.2, 27.0){\scriptsize{\rotatebox{90}{axe}}}
	\put(14.9, 25.7){\scriptsize{\rotatebox{90}{guitar}}}
	\put(16.9, 23.3){\scriptsize{\rotatebox{90}{motorbike}}}
	\put(18.8, 24.2){\scriptsize{\rotatebox{90}{babycrib}}}
	\put(20.8, 22.6){\scriptsize{\rotatebox{90}{billiardtable}}}
	\put(22.8, 23.2){\scriptsize{\rotatebox{90}{mushroom}}}
		\put(24.8, 23.6){\scriptsize{\rotatebox{90}{coffeecup}}}
		\put(26.8, 25.4){\scriptsize{\rotatebox{90}{mouse}}}
		\put(28.5, 26.5){\scriptsize{\rotatebox{90}{train}}}
		\put(30.5, 25.2){\scriptsize{\rotatebox{90}{helmet}}}
		\put(32.4, 25.8){\scriptsize{\rotatebox{90}{lizard}}}
		\put(34.4, 25.1){\scriptsize{\rotatebox{90}{banana}}}
		\put(36.2, 24.4){\scriptsize{\rotatebox{90}{ladybird}}}
		\put(38.2, 26.7){\scriptsize{\rotatebox{90}{bird}}}
		\put(40.1, 27.2){\scriptsize{\rotatebox{90}{cat}}}
		\put(42.1, 23.2){\scriptsize{\rotatebox{90}{goldenfish}}}
		\put(44.1, 26.1){\scriptsize{\rotatebox{90}{viola}}}
		\put(45.9, 26.4){\scriptsize{\rotatebox{90}{tiger}}}
		\put(47.9, 25.6){\scriptsize{\rotatebox{90}{chook}}}
	\put(49.9, 23.2){\scriptsize{\rotatebox{90}{mobulidae}}}
	\put(51.8, 26.4){\scriptsize{\rotatebox{90}{snail}}}
	\put(53.8, 26.6){\scriptsize{\rotatebox{90}{sofa}}}
	\put(55.8, 26.3){\scriptsize{\rotatebox{90}{bowl}}}
	\put(57.7, 26.6){\scriptsize{\rotatebox{90}{boat}}}
	\put(59.6, 26.0){\scriptsize{\rotatebox{90}{horse}}}
	\put(61.6, 25.5){\scriptsize{\rotatebox{90}{lemon}}}
	\put(63.6, 25.8){\scriptsize{\rotatebox{90}{snake}}}
	\put(65.4, 23.7){\scriptsize{\rotatebox{90}{aeroplane}}}
	\put(67.4, 26.7){\scriptsize{\rotatebox{90}{frog}}}
	\put(69.4, 25.8){\scriptsize{\rotatebox{90}{rabbit}}}
	\put(71.2, 26.1){\scriptsize{\rotatebox{90}{turtle}}}
	\put(73.2, 25.3){\scriptsize{\rotatebox{90}{pepper}}}
	\put(75.0, 24.6){\scriptsize{\rotatebox{90}{penguin}}}
	\put(77.0, 25.8){\scriptsize{\rotatebox{90}{camel}}}
	\put(78.8, 27.2){\scriptsize{\rotatebox{90}{pig}}}
	\put(80.8, 24.7){\scriptsize{\rotatebox{90}{monkey}}}
	\put(82.8, 25.1){\scriptsize{\rotatebox{90}{starfish}}}
	\put(84.8, 25.1){\scriptsize{\rotatebox{90}{sealion}}}
	\put(86.8, 27.2){\scriptsize{\rotatebox{90}{car}}}
	\put(88.6, 26.8){\scriptsize{\rotatebox{90}{dog}}}
	\put(90.6, 26.7){\scriptsize{\rotatebox{90}{deer}}}
	\put(92.4, 26.0){\scriptsize{\rotatebox{90}{apple}}}
	\put(94.4, 26.8){\scriptsize{\rotatebox{90}{cow}}}
	\put(96.2, 24.4){\scriptsize{\rotatebox{90}{butterfly}}}
	\put(98.2, 26.6){\scriptsize{\rotatebox{90}{bear}}}
	\put(0.7,29.2){\scriptsize{0.5}}
	\put(0.7,32.5){\scriptsize{0.6}}
	\put(0.7,35.8){\scriptsize{0.7}}
	\put(0.7,39.1){\scriptsize{0.8}}
	\put(0.7,42.4){\scriptsize{0.9}}
	\put(0.7,45.7){\scriptsize{1.0}}
	\end{overpic}
	\caption{\textbf{The performance of our model on different categories on \textit{CoSal2015} in terms of S-measure.}
	We show the prediction results of the worst group ``hammer'' and the best group ``bear''.
	These three rows from top to bottom correspond to the input image, ground truth, and our prediction.
	}\label{fig:cate}
\end{figure*}
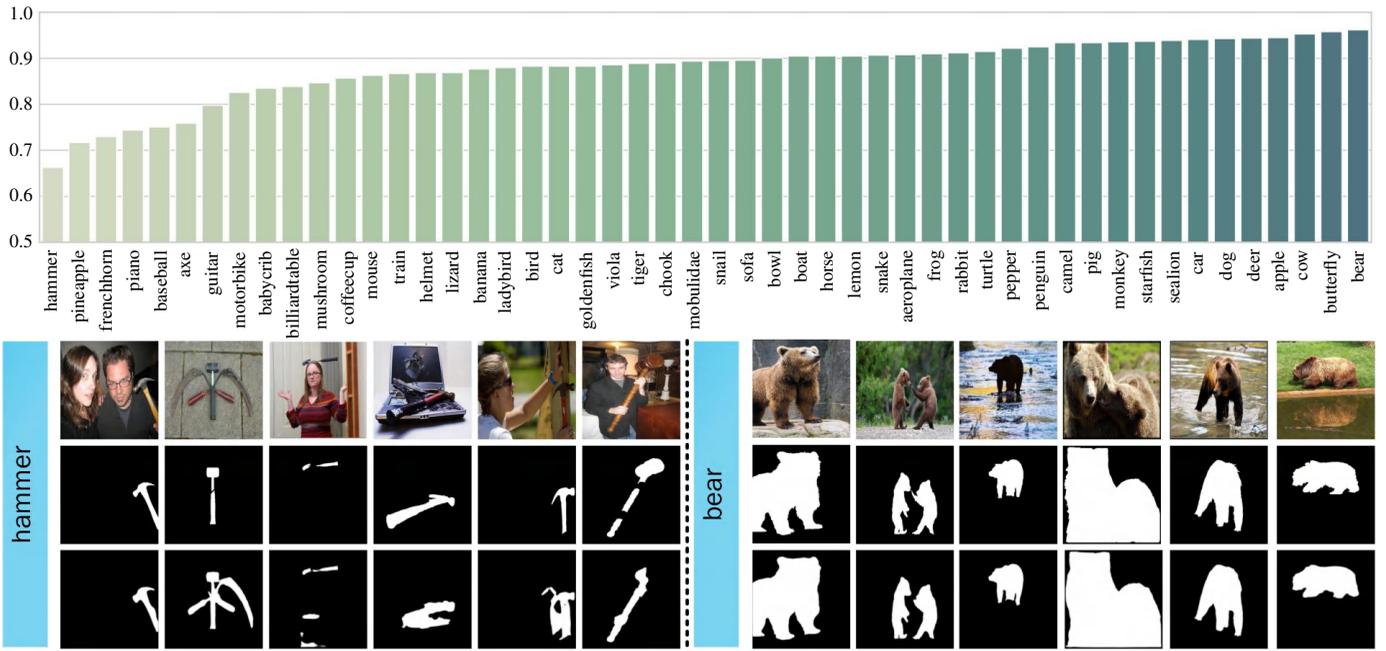

\begin{table}[t]
	\tablestyle{0.5mm}{1.3}
	\caption{
		\textbf{Influence of the number of searched sparse positions in PCS} on the \textit{CoSal2015} and \textit{CoSOD3k} datasets. 
		$K$ denotes the number of embeddings in a co-representation.
		$K^*$ is the used value we manually choose.
		$K^\text{NAS}$ is the value found by Network Architecture Search.
	}
	\begin{tabular}{l|ccc|ccc|ccc}
		\hline 
		\multirow{2}{*}{} & \multicolumn{3}{c|}{CoCA~\cite{zhang2020gicd}} & \multicolumn{3}{c|}{CoSal2015~\cite{zhang2016CoSal}} 
		& \multicolumn{3}{c}{CoSOD3k~\cite{fan2020taking}}  \\
		          & $F_{\max}\uparrow$ & $S_{\alpha}\uparrow$ & $E_{\xi}\uparrow$ 
		          & $F_{\max}\uparrow$ & $S_{\alpha}\uparrow$ & $E_{\xi}\uparrow$
		         & $F_{\max}\uparrow$ & $S_{\alpha}\uparrow$ & $E_{\xi}\uparrow$ \\
		\hline
%		\rowcolor{gray!10}
		$K$ = 16                           
		 & 0.491 & 0.635 & 0.650 & 0.865 & 0.864 & 0.900 & 0.758 & 0.791 & 0.827 \\
		$K$ = 24                           
		 & 0.542 & 0.678 & 0.708 & 0.880 & 0.871 & 0.909 & 0.788 & 0.815 & 0.856 \\
	    $K^{*}$ = 32              
	     & 0.551 & 0.686 & 0.715 & 0.885 & 0.875 & 0.913 & 0.798 & 0.820 & 0.862  \\
	    $K$ = 48                           
		 & 0.556 & 0.689 & 0.724 & 0.889 & 0.873 & 0.913 & 0.793 & 0.815 & 0.860 \\
		$K$ = 56                           
		 & 0.528 & 0.668 & 0.692 & 0.877 & 0.872 & 0.907 & 0.780 & 0.809 & 0.847 \\
%		\rowcolor{gray!10}
		$K$ = 64  
		 & 0.504 & 0.650 & 0.668 & 0.866 & 0.862 & 0.900 & 0.767 & 0.799 & 0.837  \\ 
		\hline
		$K^\text{NAS}$ = 45   
		 & 0.575 & 0.703 & 0.741 & 0.888 & 0.877 & 0.915 & 0.801 & 0.825 & 0.866  \\
		 \hline
	\end{tabular}

	\label{tab:K}
	\vspace{-5pt}
\end{table}

\subsection{Ablation Study} \label{ablations}
In \tabref{tab:ABL}, we validate the effectiveness of our pure co-representation search~(PCS) and recurrent proxy purification~(RPP).
Including SOD head increases training Dataset and 2.9 MB model parameters.
For better comparsion, we also provide the performance of baseline model with the SOD head.
Without PCS, our method losses the ability to detect co-salient objects, so in the ablation study we utilize correlation maps produced by calculating the cosine similarity between the proxy and features like ICNet~\cite{Jin2020ICNet} and GCoNet~\cite{fan2021GCoNet} in order to validate that our PCS is better than directly using the proxy as co-representation.

\myPara{Effectiveness of PCS}
PCS is designed to search multiple embeddings belonging to co-salient objects as co-representation to precisely explore co-saliency information. 
Compared with other methods, our PCS generates co-representation with less distracting noise.
With pure co-representation, our method can accurately localize the co-salient objects and finally separate them from the other parts of images.
\tabref{tab:ABL} shows that compared with ``baseline + SOD'' our PCS brings 3.2\%, 4.7\% and 6.9\% improvements in terms of S-measure on \textit{CoCA}~\cite{zhang2020gicd},  \textit{CoSOD3k}~\cite{fan2020taking} and \textit{CoSal2015}~\cite{zhang2016CoSal} datasets.
On condition that the number of parameters in our method is not larger than that in the ``baseline + SOD'', this result demonstrates the effectiveness of our PCS.
Besides, the co-representation proxy can be directly used as co-representation to explore co-salient objects, but compared with ``baseline + SOD + proxy'', our pure co-representation search brings 8.9\%, 5.6\% and 3.3\% improvement in terms of S-measure on \textit{CoCA}~\cite{zhang2020gicd}, \textit{CoSOD3k}~\cite{fan2020taking} and \textit{CoSal2015}~\cite{zhang2016CoSal} datasets correspondingly.
The improvment benefits from the advantages analysed in \secref{sec:pcs}.

The co-representation of CoRP is made up of \textit{K} embeddings, most of which belong to co-salient objects.
Our co-representation will contain more noise if the \textit{K} is too large, and the diversity of co-salient information will decrease if the \textit{K} is too small.
In \tabref{tab:K}, we design an experiment where the hyperparameter \textit{K} is set  to different values in order to find the proper \textit{K} for our CoRP.
According to the performance on \textit{CoSal2015}, \textit{CoCA} and \textit{CoSOD3k} datasets with different \textit{K}, we set \textit{K} to 32 in all the other experiments.

Further, we employ \textit{Network Architecture Search} (NAS) to find the optimal $K$.
Specifically, 
we follow the NAS method \cite{guo2020single} to implement this search process.
The last row of \tabref{tab:K} shows that we use NAS to find a better $K$ for the three benchmark Co-SOD datasets.
Compared with the randomly set $K$,
the $K$ searched by NAS brings obvious improvement on \textit{CoCA} dataset.

\begin{table}[t]
	\tablestyle{0.5mm}{1.45}
	\caption{
		\textbf{Performance of CoRP along with iterations} on the \textit{CoCA} and \textit{CoSOD3k} datasets.
		\textit{T} denotes the times of iteration.
	}
	\begin{tabular}{c|c|ccc|ccc|ccc}
		\hline 
		\multirow{2}{*}{}  &  
		& \multicolumn{3}{c|}{CoCA~\cite{zhang2020gicd}}
		&\multicolumn{3}{c|}{CoSal2015~\cite{zhang2016CoSal}} & \multicolumn{3}{c}{CoSOD3k~\cite{fan2020taking}}   \\ & \textit{FPS}
		         & $F_{\max}\uparrow$ & $S_{\alpha}\uparrow$ & $E_{\xi}\uparrow$  
		         & $F_{\max}\uparrow$ & $S_{\alpha}\uparrow$ & $E_{\xi}\uparrow$ & $F_{\max}\uparrow$ & $S_{\alpha}\uparrow$ & $E_{\xi}\uparrow$   \\
		\hline
%		\rowcolor{gray!10}
	    $T$ = 1  & 62.5                          
		& 0.488 & 0.640 & 0.658 & 0.874 & 0.869 & 0.906 & 0.764 & 0.800 & 0.839  \\
	    $T$ = 2  & 56.6              
	    & 0.532 & 0.672 & 0.699 & 0.883 & 0.873 & 0.901  & 0.790 & 0.815 & 0.857 \\
%		\rowcolor{gray!10}
		$T$ = 3  & 45.3  
		& 0.551 & 0.686 & 0.715 & 0.885 & 0.875 & 0.913  & 0.798  & 0.820 & 0.862\\ 
		$T$ = 4  & 39.5  
		& 0.561 & 0.691 & 0.722 & 0.887 & 0.875 & 0.914 & 0.800 & 0.821 & 0.863  \\
		$T$ = 5  & 34.8
		& 0.565 & 0.693 & 0.725 & 0.887 & 0.875 & 0.914 & 0.800 & 0.821 & 0.864  \\
		$T$ = 6  & 30.1  
		& 0.567 & 0.693 & 0.727 & 0.887 & 0.875 & 0.914 & 0.800 & 0.821 & 0.864  \\
		\hline
	\end{tabular}
	\vspace{-5pt}
	\label{tab:RPP_iteration}
	% \vspace{-20pt}
\end{table}

\begin{table}[t!]
	\tablestyle{0.3mm}{1.45}
	\caption{
		\textbf{Results of our method with different batch size settings.} 
	    $n_{train}$ and
        $n_{test}$ denote training and test input size, respectively.
        ``all'' denotes all images of a category. 
	}
	\begin{tabular}{cc|ccc|ccc|ccc}
		\hline 
		\multirow{2}{*}{}  &  
		& \multicolumn{3}{c|}{CoCA~\cite{zhang2020gicd}}
		&\multicolumn{3}{c|}{CoSal2015~\cite{zhang2016CoSal}} & \multicolumn{3}{c}{CoSOD3k~\cite{fan2020taking}}   \\
		$n_{train}$ & $n_{test}$
		         & $F_{\max}\uparrow$ & $S_{\alpha}\uparrow$ & $E_{\xi}\uparrow$  
		         & $F_{\max}\uparrow$ & $S_{\alpha}\uparrow$ & $E_{\xi}\uparrow$ & $F_{\max}\uparrow$ & $S_{\alpha}\uparrow$ & $E_{\xi}\uparrow$   \\
		\hline
%		\rowcolor{gray!10}
	    5  & 5                          
		& 0.540 & 0.674 & 0.702 & 0.878 & 0.872 & 0.906 & 0.791 & 0.810 & 0.854  \\
	    5  & 10              
	    & 0.549 & 0.678 & 0.707 & 0.881 & 0.873 & 0.910  & 0.795 & 0.815 & 0.856 \\
%		\rowcolor{gray!10}
		5  & 15  
		& 0.549 & 0.681 & 0.712 & 0.882 & 0.874 & 0.912  & 0.794  & 0.819 & 0.860 \\ \hline
		10  & 5  
		& 0.541 & 0.676 & 0.704 & 0.879 & 0.871 & 0.907 & 0.792 & 0.813 & 0.853  \\
		10  & 10
		& 0.548 & 0.683 & 0.710 & 0.883 & 0.874 & 0.914 & 0.797 & 0.818 & 0.860  \\
		10  & 15  
		& 0.550 & 0.686 & 0.713 & 0.885 & 0.875 & 0.912 & 0.798 & 0.819 & 0.859  \\ \hline
		10  & all  
		& 0.551 & 0.686 & 0.715 & 0.885 & 0.875 & 0.913  & 0.798  & 0.820 & 0.862\\
		\hline
	\end{tabular}
  \vspace{-5pt}
	\label{tab:number}
	% \vspace{-20pt}
\end{table}

\myPara{Effectiveness of RPP}
We employ RPP to purify the co-representation proxy, which is used for searching the pure co-representation.
Many visualization results and statistics demonstrate the effectiveness of the iterative process.
In \figref{fig:iters},
based on iterations,
the distribution of co-representation proxy is closer to the ground-truth one.
Thanks to the better co-representation proxies, 
in \tabref{tab:percentage}, 
co-representations consist of a higher percentage of embeddings belonging to co-salient regions after each iteration.
\figref{fig:recurrent} intuitively illustrates that our method gradually eliminates the error predictions. 
\tabref{tab:ABL} and \tabref{tab:Results} show that RPP improves our performance and enables our CoRP to achieve SOTA methods. 

We investigate the influence of iteration times on model performance in \tabref{tab:RPP_iteration}.
With the progress of the iterative process, 
the performance of our model is getting better and better, but at the same time, the increase is gradually decreasing.
The effectiveness of RPP is significant from $T$=1 to $T$=2, which brings 3.2\%, 1.5\% and 0.4\% improvements in terms of S-measure on \textit{CoCA}~\cite{zhang2020gicd}, \textit{CoSOD3k}~\cite{fan2020taking} and \textit{CoSal2015}~\cite{zhang2016CoSal} correspondingly.
As there are many distracting foreground objects in \textit{CoCA}~\cite{zhang2020gicd},
the improvement fully reveals the process of purifying co-representation.
Although the fourth and fifth iteration still have gains in performance, in all other experiments, 
we let our CoRP end in the third iteration ($T$ = 3) considering the balance between the performance of our method and the computational cost.

\myPara{Impacts of the input size}
In \tabref{tab:number}, we provide the performances of our model with the different number of samples in a group during training and testing.
Increasing the number of samples in training can make the training process more stable.
But restricted by GPU memory, we set the number into 10 in our final model.
Meanwhile,
compared with the training stage,
a larger input size in testing brings more obvious improvement.

\myPara{Impacts of different groups}
In \figref{fig:cate},
we show both qualitative and quantitative results on different categories.
We find that our performance is relatively stable.
The S-measure of most categories are over 0.850.
For the best ``bear'' group,
the segmentation results are very close to the ground truths.
However, on few extremely ``difficult'' groups,
the performance can drop.
For the ``hammer'' group, 
the hammers are small and relatively hidden.
We still successfully locate the hammers,
but the segmentation details are not perfect.

\begin{table}[t]
	\tablestyle{0.7mm}{1.3}
	\caption{
		\revise{\textbf{The performance of our model with different $\alpha$ and $\beta$.}} 
	}
	\begin{tabular}{cc|ccc|ccc|ccc}
		\hline 
		%[-0.1cm]
		%[-0.3ex]
		\multirow{2}{*}{} & & \multicolumn{3}{c|}{\revise{CoCA~\cite{zhang2020gicd}}} & \multicolumn{3}{c|}{\revise{CoSal2015~\cite{zhang2016CoSal}}}  & \multicolumn{3}{c}{\revise{CoSOD3k~\cite{fan2020taking}}}   \\
		           \revise{$\alpha$} & \revise{$\beta$}  & \revise{$F_{\max}\uparrow$} & \revise{$S_{\alpha}\uparrow$} & \revise{$E_{\xi}\uparrow$}    & \revise{$F_{\max}\uparrow$} & \revise{$S_{\alpha}\uparrow$} & \revise{$E_{\xi}\uparrow$} 
		         & \revise{$F_{\max}\uparrow$} & \revise{$S_{\alpha}\uparrow$} & \revise{$E_{\xi}\uparrow$} \\
		%[-0.1cm]
		\hline
%		\rowcolor{gray!10}
		\revise{0.2}  & \revise{0.8} 
		 & \revise{0.541} & \revise{0.677} & \revise{0.716}  & \revise{0.881} & \revise{0.867} & \revise{0.905}  & \revise{0.792} & \revise{0.814} & \revise{0.859}  \\
		\revise{0.4}  & \revise{0.6}                          
		 & \revise{0.542} & \revise{0.677} & \revise{0.703} & \revise{0.883} & \revise{0.874} & \revise{0.910}  & \revise{0.784} & \revise{0.813} & \revise{0.853}  \\
	    \revise{0.5} & \revise{0.5}             
	     & \revise{0.540} & \revise{0.674} & \revise{0.700}  & \revise{0.883} & \revise{0.874} & \revise{0.910}   & \revise{0.788} & \revise{0.817} & \revise{0.856}  \\
%		\rowcolor{gray!10}
		\revise{0.6}  & \revise{0.4}  
		 & \revise{0.547} & \revise{0.680} & \revise{0.717}  & \revise{0.886} & \revise{0.874} & \revise{0.911}  & \revise{0.790} & \revise{0.815} & \revise{0.855}   \\  
		\revise{0.8} & \revise{0.2}     
		 & \revise{0.551} & \revise{0.686} & \revise{0.715}  & \revise{0.885} & \revise{0.875} & \revise{0.913} & \revise{0.798} & \revise{0.820} & \revise{0.862}  \\
		\hline
	\end{tabular}
	\label{tab:ab}
	% \vspace{-20pt}
\end{table}

\myPara{\revise{Impacts of the hyper parameters $\alpha$ and $\beta$}}
\revise{
In our loss function, $\alpha$ and $\beta$ respectively control the supervisions for co-salient maps and salient maps.
To explore their impacts on the results, 
we set $\alpha$ and $\beta$ to different values and show the corresponding qualitative results in \tabref{tab:ab}.
We can see that setting $\alpha > \beta$ can generate better results and we finally set $\alpha=0.8$, $\beta=0.2$.}

\begin{table*}[t]
	\tablestyle{4.3mm}{1.3}
	\caption{
		\textbf{Quantitative comparisons} of \textit{Precision} ($\mathcal{P}$) and \textit{Jaccard} ($\mathcal{J}$) by our CoRP and other co-segmentation methods on the co-segmentation
        benchmark dataset Internet. 
	}
	\begin{tabular}{c|cc|cc|cc|cc}
		\hline 
		\multirow{2}{*}{Internet Dataset} &  \multicolumn{2}{c|}{Airplane} & \multicolumn{2}{c|}{{Car}}  & \multicolumn{2}{c|}{Horse} & \multicolumn{2}{c}{Average}\\
		          & $\mathcal{P}(\%)\uparrow$ & $\mathcal{J}(\%)\uparrow$ & $\mathcal{P}(\%)\uparrow$ & $\mathcal{J}(\%)\uparrow$ &  $\mathcal{P}(\%)\uparrow$ & $\mathcal{J}(\%)\uparrow$ & $\mathcal{P}(\%)\uparrow$ & $\mathcal{J}(\%)\uparrow$\\
		%[-0.1cm]
		\hline
%		\rowcolor{gray!10}
		Jerripothula \etal{} \cite{jerripothula2016image}
		& 90.5 & 61.0 & 88.0 & 71.0 & 88.3 & 60.0 & 88.9 & 64.0   \\
		Li \etal{} \cite{li2018deep}           
		& 94.1 & 65.4 & 93.9 & 82.8 & 92.4 & 69.4 & 93.5 & 72.5    \\
	    Chen \etal{} \cite{chen2020show}         
	    & 94.1  & 65.0 & 94.0 & 82.0 & 92.2 & 63.0 & 93.4 & 70.0   \\
%		\rowcolor{gray!10}
		Zhang \etal{}\cite{zhang2020deep} 
		& 94.6 & 66.7 & 89.7 & 68.1 & 93.2 & 66.2 & 92.5 & 67.0    \\ \hline
		Ours   
		& 94.4 & 83.0 & 94.1 & 93.0 & 93.9 & 77.0 & 94.1 & 84.3   \\
		\hline
	\end{tabular}
	\label{tab:Internet}
\end{table*}

\begin{table*}[ht]
	\tablestyle{3mm}{1.3}
  \caption{
		\textbf{Quantitative comparisons} of \textit{Jaccard} ($\mathcal{J}$) by our CoRP and other co-segmentation methods on the co-segmentation dataset iCoseg. 
	}
\begin{tabular}{c|ccccccccc} \hline
  iCoseg Dataset & Average$\mathcal{J}(\%)\uparrow$ & bear2 & brownbear & cheetah & elephant & helicopter & hotballoon & panda1 & panda2 \\ \hline
  Jerripothula \etal \cite{jerripothula2016image} 
	& 70.4 & 67.5 & 72.5 & 78.0 & 79.9 & 80.0 & 80.2 & 72.2 & 61.4 \\ \hline
  Li \etal \cite{li2018deep} 
	& 84.2 & 88.3 & 92.0 & 68.8 & 84.6 & 79.0 & 91.7 & 82.6 & 86.7 \\ \hline
  Chen \etal \cite{chen2018semantic} 
	& 86.0 & 88.3 & 91.5 & 71.3 & 84.4 & 76.5 & 94.0 & 91.8 & 90.3 \\ \hline
  Zhang \etal \cite{zhang2020deep} 
	& 88.0 & 87.4 & 90.3 & 84.9 & 90.6 & 76.6 & 94.1 & 90.6 & 87.5 \\ \hline
  Ours 
	& 90.5 & 91.6 & 92.8 & 90.1 & 91.2 & 79.3 & 95.6 & 93.9 & 89.5 \\ \hline
\end{tabular}
\label{tab:iCoseg}
\vspace{-5pt}
\end{table*}

\begin{table}[t]
	\tablestyle{7mm}{1.3}
	\caption{
		\textbf{Quantitative comparisons} of \textit{Precision} ($\mathcal{P}$) and \textit{Jaccard} ($\mathcal{J}$) by our CoRP and other methods on the 
		co-segmentation benchmark co-segmentation dataset MSRC. 
	}
	\begin{tabular}{c|c|c} \hline 
		MSRC Dataset & $\mathcal{P}(\%)\uparrow$ & $\mathcal{J}(\%)\uparrow$ \\
		\hline
		Mukherjee \etal \cite{mukherjee2018object} & 84.0 & 67.0   \\
		Li \etal \cite{li2018deep} & 92.4 & 79.9     \\
	  Chen \etal \cite{chen2018semantic} & 95.2  & 77.7   \\
		Zhang \etal \cite{zhang2020deep} 		& 94.3 & 79.4    \\ \hline
		Ours   		& 96.0 & 83.1   \\
		\hline
	\end{tabular}
	\label{tab:MSRC}
\end{table}

\begin{figure}[ht!]
	\centering
	\includegraphics[width=\linewidth]{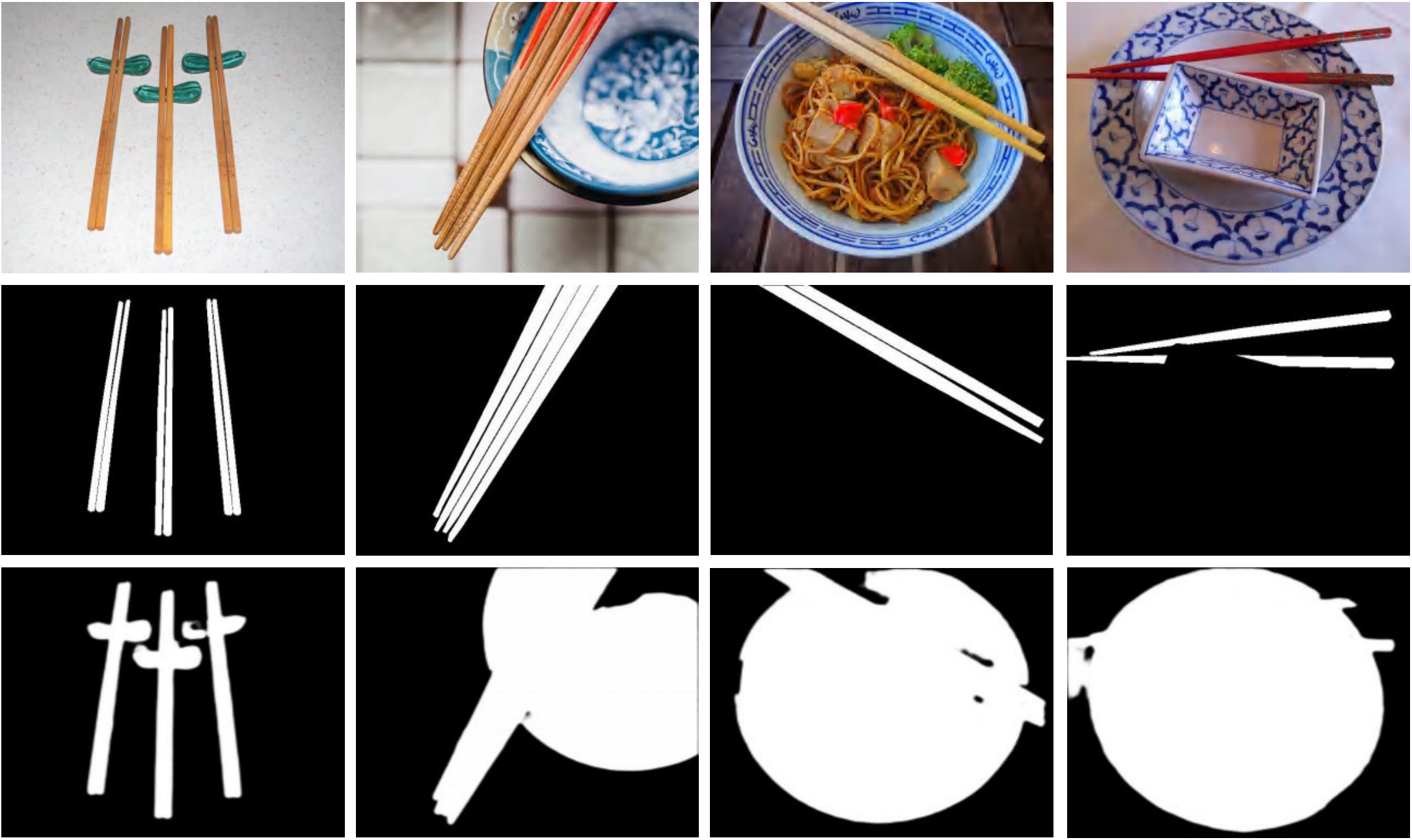} \\
	\vspace{-5pt}
	\caption{\revise{\textbf{Failure cases of our method.}
	  From top to bottom are images, ground truths and our predictions.
	  The chopsticks are the co-salient objects of this group.}
	}\label{fig:failure}
\end{figure}

\revise{\subsection{Failure Case}
For some extremely challenging scenarios,
our method cannot successfully segment the co-salient objects.
Take the images in \figref{fig:failure} as an example,
the chopsticks are the co-salient objects of this group.
However,
the salient object bowls appear in most of the images.
Meanwhile,
chopsticks are smaller objects compared with bowls.
and the bowls are more salient than the chopsticks.
In this case,
it is difficult to eliminate noise from the bowls and extract a pure co-representation,
so that the predictions are inaccurate.}

\subsection{Extention to Co-Segmentation}

To validate the universal effectiveness of our CoRP,
we extend our method to the field of image co-segmentation.
In fact, image co-segmentation is similar to Co-SOD.
Both of the two tasks aim at segmenting the common objects among a group of relevant images.
The difference is that image co-segmentation does not need the co-objects to be salient.
Our method, 
which aims at solving complex scenarios in Co-SOD,
is also very effective in co-segmentation.

We also qualitatively compare our CoRP with popular co-segmentation methods on the three co-segmentation benchmark datasets,
including \textit{MSRC} dataset \cite{shotton2006textonboost}, \textit{iCoseg} dataset \cite{batra2010icoseg}, and \textit{Internet} dataset \cite{rubinstein2013unsupervised}.
We report the performances of the co-segmentation models using two wildly used metrics: \textit{Precision} and \textit{Jaccard}.

\begin{figure}[t]
	\centering
	\includegraphics[width=\linewidth]{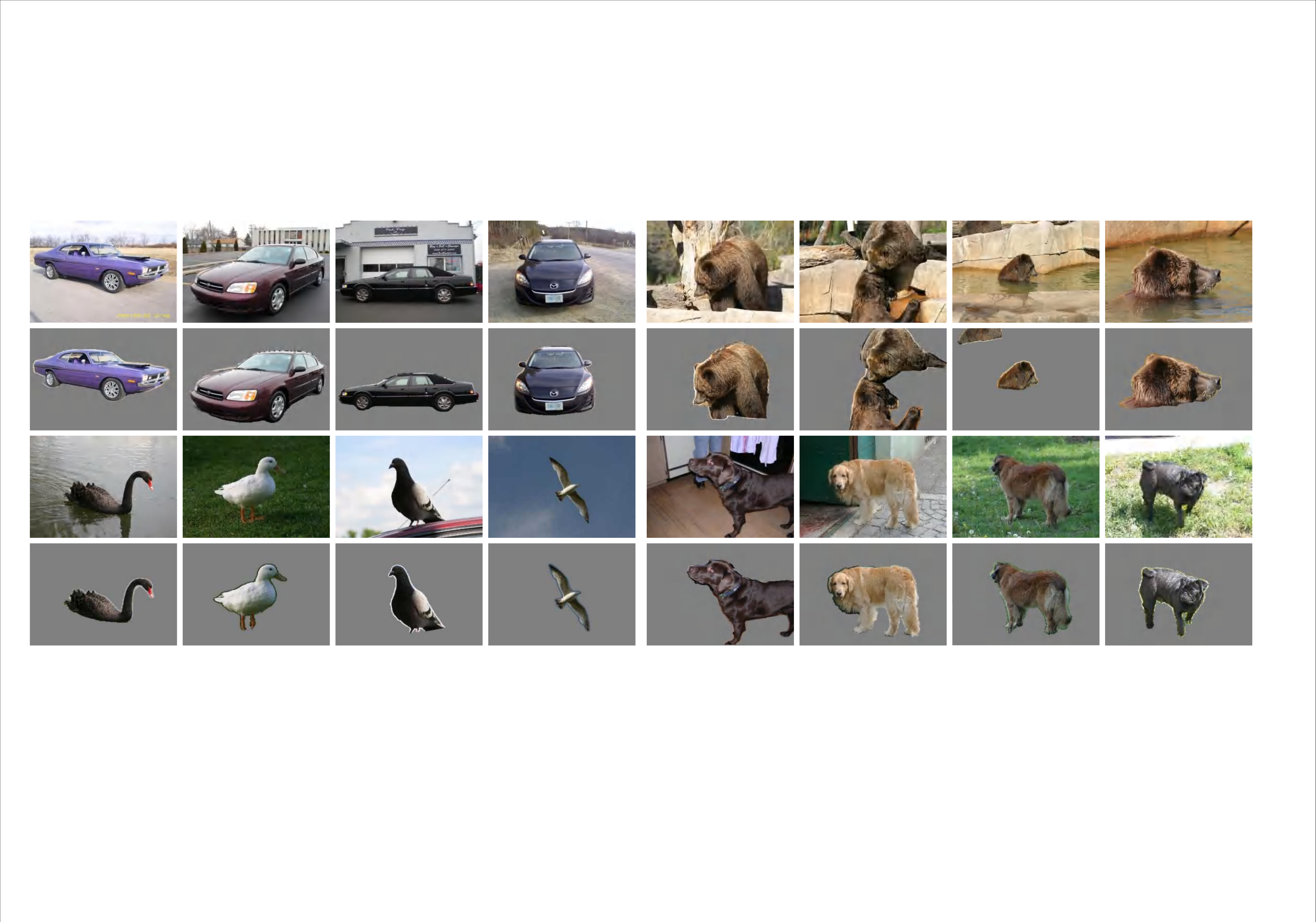} \\
	\vspace{-5pt}
	\caption{\textbf{Our qualitative results on the co-segmentation datasets.}
	}\label{fig:coseg}
\end{figure}

The \tabref{tab:Internet} shows that on \textit{Internet} dataset,
our outperform the other methods across all categories and all metrics.
Our improvement in terms of \textit{Jaccard} is very obvious.
Compared with the second-best method, 
our method brings 17.3\% improvement on the overall dataset.
In terms of \textit{Precision}, 
our method also brings 0.6\% improvement.
In \tabref{tab:iCoseg},
we compare our method with others in terms of \textit{Jaccard}.
Once again, we outperform other methods in all categories.
On the overall \textit{iCoseg} dataset,
our method brings 2.5\% improvement compared with the second-best method.
Further,
the \tabref{tab:MSRC} reveals that on \textit{MSRC}
dataset,
our method respectively bring 1.7\% and 3.7\% performance improvement in term of \textit{Precision} and \textit{Jaccard}.
In \figref{fig:coseg},
we present two groups (car and bird) of our predictions.
It shows that our method segments the target objects with high accuracy.

\section{Conclusions}
In this paper, we observe that current Co-SOD methods have not paid due attention to the noise in co-representation which is crucial for Co-SOD task,
so that their performance suffer from the complex foreground objects' interference.
To overcome this shortcoming,
we focus on eliminating the distracting information in the co-representation and propose an effective method (CoRP).
Our CoRP works iteratively with two collaborative strategies (PCS and RPP), which aim at suppressing the noise in the co-representation to acquire more accurate prediction.
In a nutshell, 
PCS is designed for searching pure co-representation from the sparse locations belonging to co-salient objects with the help of co-representation proxy provided by RPP,
then our PCS feedbacks newly predicted co-saliency maps to RPP.
Again, RPP uses the newly predictions to calculate a co-representation proxy with less distracting information in turn.
In this way,
the co-representation and prediction are utilized to improve each other in an iterative process.
Substantial visualization results and experimental analysis demonstrate our contributions.
Our CoRP achieves SOTA results on three challenging datasets.

\myPara{Acknowledgement}
This research was supported by the NSFC (62176130)
and the Fundamental Research Funds for the Central Universities 
(Nankai University, NO. 63223050).

% In this paper, we propose a Co-Representation Purification (CoRP) method for Co-SOD.
% We notice that the distracting information in co-representation has serious influence on the prediction of co-saliency maps.
% To overcome the drawback, 
% we firstly purify our co-representation by searching multiple representations belonging to co-salient objects as co-representation which is used as guide to generate prediction.
% Then we iteratively utilize the prediction prediction to reduce the distracting information in the co-representation and get better prediction.
% The two mechanisms we proposed work collaboratively to extract purified co-representation and finally make our method achieve a SOTA performance on three benchmark datasets.
% Through comprehensive visualization experiments and ablation studies we also validate the rationality of our statement and effectiveness of purifying the co-representation.

\bibliographystyle{IEEEtran}
\bibliography{egbib}

\newcommand{\Auther}[1]{ \includegraphics[width=1in,keepaspectratio]{figures/Authors/#1}}

% \Author{ziyuezhu}{Zi-Yue Zhu}
% {is currently a master student from the College of Computer Science at Nankai University, under the supervision of Prof. Ming-Ming Cheng. His research interests include deep learning and computer vision.}
% \vspace{-0.95in}
\begin{IEEEbiography}[\Auther{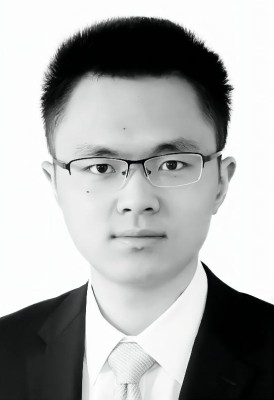}]
{Zi-Yue Zhu} is currently a master student from the College of Computer Science at Nankai University, under the supervision of Prof. Ming-Ming Cheng. His research interests include deep learning and computer vision.
\end{IEEEbiography}
\vspace{-.4in}

\begin{IEEEbiography}[\Auther{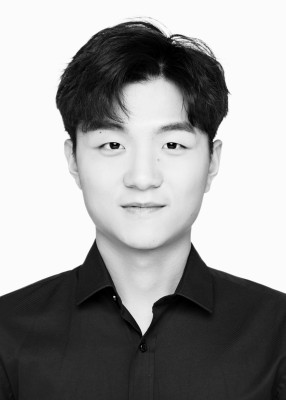}]
{Zhao Zhang} is currently a researcher at SenseTime Group LTD. 
He received a Master's degree at Nankai University under the supervision of Prof. Ming-Ming Cheng, and a Bachelor's degree at Yangzhou University. His research interests mainly focus on image processing, computer vision, and deep learning.
\end{IEEEbiography}
\vspace{-.4in}

\begin{IEEEbiography}[\Auther{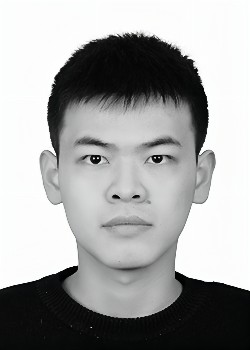}]
{Zheng Lin} is currently a Ph.D. candidate with College of Computer Science, Nankai University, under the supervision of Prof. Ming-Ming Cheng. His research interests include deep learning, computer graphics, and computer vision.
\end{IEEEbiography}
\vspace{-.4in}

\begin{IEEEbiography}[\Auther{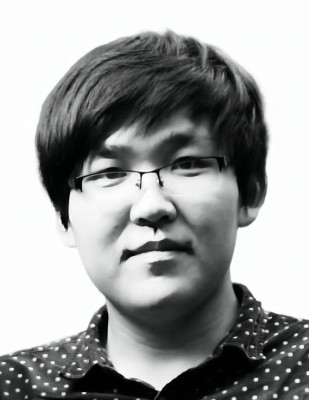}]
{Xing Sun} is currently a Principal researcher in Tencent YoutuLab. Before that, he received his Ph.D. degree from The University of Hong Kong in 2016. His research interests include image processing, machine learning, and computer vision.
\end{IEEEbiography}
\vspace{-.4in}

\begin{IEEEbiography}[\Auther{cmm.jpg}]
{Ming-Ming Cheng} received his PhD degree
from Tsinghua University in 2012. 
Then he did 2
years research fellow, 
with Prof. Philip Torr in Oxford. 
He is now a professor at Nankai University,
leading the Media Computing Lab. 
His research interests includes computer graphics, 
computer vision, 
and image processing. 
He received research awards including ACM China Rising Star
Award, 
IBM Global SUR Award, CCF-Intel Young
Faculty Researcher Program, \etc.
\end{IEEEbiography}

\vfill

\end{document}